\documentclass{article}
\UseRawInputEncoding
\usepackage[T1]{fontenc}
\usepackage[utf8]{inputenc}
\usepackage{microtype}
\usepackage{inconsolata}

\usepackage{graphicx} % Required for inserting images
\usepackage{booktabs}
\usepackage{lipsum}
\usepackage{multicol}
\usepackage{multirow}
\usepackage{color,xcolor,colortbl}
\usepackage{enumitem}
\usepackage{amssymb} 

\usepackage{times}
\usepackage{latexsym}
\usepackage{amsmath} 

\usepackage{subcaption}
\usepackage{hyperref}

\newcommand\blfootnote[1]{%
  \begingroup
  \renewcommand\thefootnote{}\footnote{#1}%
  \addtocounter{footnote}{-1}%
  \endgroup
}

\title{Uncertainty-aware abstention in medical diagnosis based on medical texts}
\author{Artem Vazhentsev${}^2$, Ivan Sviridov${}^3$, Alvard Barseghyan${}^4$, \\ Gleb Kuzmin${}^2$, Alexander Panchenko${}^2$, Aleksandr Nesterov${}^2$, \\ Artem Shelmanov${}^1$ and Maxim Panov${}^1$\blfootnote{${}^1$ MBZUAI, Abu Dhabi, UAE. ${}^2$ AIRI, Moscow, Russia. ${}^3$ Sber AI Lab, Moscow, Russia. ${}^4$ YerevaNN, Yerevan, Armenia. e-mail: vazhetsev@airi.net; wchhiaarid@gmail.com; maxim.panov@mbzuai.ac.ae}}
\date{}%{January 2025}

\newcommand{\xv}{\mathbf{x}}
\newcommand{\uv}{\mathbf{u}}
\newcommand{\XSet}{\mathcal{X}}
\newcommand{\DC}{\mathcal{D}}

\begin{document}

\maketitle

\begin{abstract}
  This study addresses the critical issue of reliability for AI-assisted medical diagnosis. We focus on the selection prediction approach that allows the diagnosis system to abstain from providing the decision if it is not confident in the diagnosis. Such selective prediction (or abstention) approaches are usually based on the modeling predictive uncertainty of machine learning models involved.

  This study explores uncertainty quantification in machine learning models for medical text analysis, addressing diverse tasks across multiple datasets. We focus on binary mortality prediction from textual data in MIMIC-III, multi-label medical code prediction using ICD-10 codes from MIMIC-IV, and multi-class classification with a private outpatient visits dataset. Additionally, we analyze mental health datasets targeting depression and anxiety detection, utilizing various text-based sources, such as essays, social media posts, and clinical descriptions.

  In addition to comparing uncertainty methods, we introduce HUQ-2, a new state-of-the-art method for enhancing reliability in selective prediction tasks. Our results provide a detailed comparison of uncertainty quantification methods. They demonstrate the effectiveness of HUQ-2 in capturing and evaluating uncertainty, paving the way for more reliable and interpretable applications in medical text analysis.
\end{abstract}

\section{Introduction}
\label{sec:intro}
  Artificial intelligence (AI) plays an important role in the classification and analysis of medical texts~\cite{Sammani2021, JUHN2020463}, offering improvements in efficiency and precision compared to traditional text processing pipelines~\cite{Lu2022}. Using AI-based solutions, healthcare professionals can automate complex decision-making processes~\cite{Comendador2014PharmabotAP, Ni2017MANDYTA} and minimize human error~\cite{pmid32269070, PMID:31068188}. These advancements hold immense potential, particularly in clinical decision support~\cite{sutton2020overview}, medical coding~\cite{Alyahya2019-gq}, and diagnostic report generation~\cite{park2024patient}.

  Classifying medical texts, a critical component of Electronic Health Records (EHR) processing pipelines, is essential for interpreting unstructured data such as clinical notes and doctor reports. These tasks are uniquely challenging due to the complexity and specialized language of medical texts, which often include domain-specific terms, abbreviations, and varied writing styles. For example, the medical code assignment problem is an inherently highly ambiguous multi-label classification task~\cite{LIU2023102662}, requiring the selection of multiple codes for a single document from extensive and highly intricate classification systems, such as the International Classification of Diseases (ICD)\footnote{ICD: \url{https://icd.who.int/browse10/2019/en}}. This complexity demands careful consideration of hierarchical and interdependent code relationships while ensuring that all relevant ones are identified accurately, making medical text classification a particularly demanding area within natural language processing~\cite{MUJTABA2019494}.
  
  Given the critical role of working with medical texts in healthcare, the accurate classification of medical documents becomes even more critical, as these systems rely on precise information extraction to generate reliable diagnoses. Misclassification can result in significant consequences, including misdiagnosis~\cite{pmid32269070}, improper treatment~\cite{Alyahya2019-gq}, or even increased mortality~\cite{PMID:31068188}. Therefore, improving the classification methods for medical texts is crucial to enhancing the performance of automatic diagnosis systems and ensuring better patient outcomes.

  To address these challenges and reduce the risks associated with errors in classification, the medical domain imposes stringent requirements on AI algorithms, particularly regarding the reliability and transparency of their decision-making processes. The need for explainability is critical, as healthcare decisions often involve high stakes, where the accuracy and trustworthiness of AI predictions are paramount. As a result, there is a strong demand for methods that ensure high predictive performance and the ability to explain how to reach conclusions. One such method involves a qualitative approach to model validation, which assesses the model's behavior systematically and interpretably. Validation workflows include statistical rigor, such as cross-validation and performance metrics, alongside clinical interpretability by testing models in real-world medical scenarios~\cite{pmid35426190}. This estimation ensures that models are evaluated for accuracy and practical utility in clinical contexts. Additionally, transparency in describing the AI model's architecture, including its training pipeline, data preprocessing, and decision rules, is essential for building trust with medical practitioners and ensuring compliance with regulatory standards~\cite{10.1093/jamia/ocaa088}.

  Nevertheless, even with robust validation methods and transparent architectures, relying solely on model predictions remains risky, as incorrect predictions can mislead doctors, causing potential harm to patients. In medical applications, where decisions can have life-altering consequences, assessing and communicating predictions' reliability is paramount. A potential solution to mitigate this risk is incorporating a degree of confidence in the model's predictions, a technique commonly called \textit{uncertainty quantification} (UQ; \cite{gawlikowski2022survey}). This research direction is crucial for medical processing systems, as it aims to enhance the reliability and safety of AI models by providing mechanisms to assess and quantify the confidence in their predictions~\cite{Kompa2021}. This capability is particularly valuable for identifying cases that require additional scrutiny, such as those involving inherent ambiguity~\cite{10.1093/jamia/ocaa269}, rare diseases~\cite{chen-etal-2023-rare}, or out-of-distribution samples~\cite{Zadorozhny2023}, which are common challenges in healthcare applications.

  \begin{figure*}[t!]
    \centering
    \includegraphics[trim={0.cm 0.cm 0.cm 0.cm},clip,width=1.\linewidth]{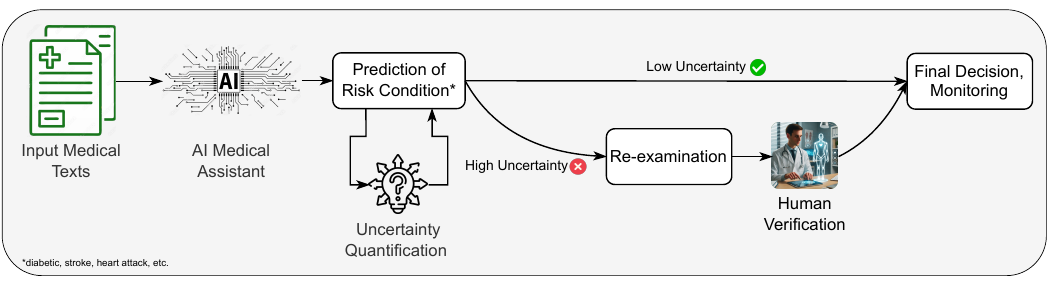}
    \caption{
    An illustration of the verification pipeline in medicine based on uncertainty quantification. The most uncertain predictions are checked additionally by medical professionals.}
    \label{fig:med_uq_pipeline}
  \end{figure*}

  Based on uncertainty quantification methods, \textit{selective prediction}~\cite{selective-classification}, also known as \textit{abstention}~\cite{xin-etal-2021-art}, has emerged to improve the reliability of machine learning models. Leveraging uncertainty quantification allows models to abstain from predictions when uncertainty is high, ensuring they make only confident predictions. This capability is crucial in medicine since selective prediction ensures that critical decisions are based only on reliable outputs, safeguarding against potential harm. For instance, in detecting critical diseases such as heart failure or diabetes~\cite{who}, uncertainty quantification helps prioritize predictions with higher confidence, particularly for rare or severe cases. We illustrate the resulting medical diagnosing pipeline powered with selective prediction in Figure~\ref{fig:med_uq_pipeline}.

  This approach enhances diagnostic accuracy by reducing overconfident errors and ensuring that ambiguous cases are flagged for further review, ultimately supporting safer and more reliable medical decision-making~\cite{li2020deep}. In other words, by integrating UQ-based methods, medical systems can enhance patient safety, build trust in AI-driven healthcare, and provide a transparent mechanism for balancing automation with human oversight~\cite{deuschel2024role}.
  
  %High-uncertainty cases can be flagged for further review by medical professionals, adding a layer of safety before making clinical decisions. This approach enhances patient safety and trust in AI systems~\cite{deuschel2024role} by providing transparent uncertainty metrics.

  In this work, we focus on selective prediction for medical diagnosis tasks based on medical textual data. While numerous recent methods have been developed for uncertainty quantification and selective prediction using texts~\cite{vazhentsev-etal-2022-uncertainty,fadeeva2023lm,vazhentsev-etal-2023-hybrid,ren2023outofdistribution}, their application in medicine remains limited~\cite{ASHFAQ2023104019, PELUSO2024104576}. This work comprehensively evaluates state-of-the-art uncertainty quantification techniques across diverse tasks, including mortality prediction, mental disorder detection, and multi-class and multi-label medical code prediction (MCP) using medical texts. In particular, we focus on modeling of \textit{epistemic}\footnote{Epistemic (model) uncertainty is a part of predictive uncertainty that is due to the lack of knowledge about the observed phenomenon and modeling it based on the finite data sample.} and \textit{aleatoric}\footnote{Aleatoric (data) uncertainty is a part of predictive uncertainty that is due to the inherent randomness in the data such as label noise or label ambiguity} uncertainty, and demonstrate the importance of both of them for the efficiency of selective prediction. Additionally, we highlight the challenges and opportunities of uncertainty quantification for multi-label classification, where selective prediction can be performed not only on the instance level but also for particular diagnoses, leading to improved performance. 
  
  The main contributions of this paper can be summarized as follows:
  \begin{enumerate}
    \item We perform the largest to date study of state-of-the-art uncertainty quantification methods in the task of selective prediction for mortality prediction, mental disorder detection, multi-class, and multi-label medical code prediction based on EHR data. Our experiments include not only public benchmarks but also a new comprehensive real-world dataset, which is essential to test methods under realistic clinical conditions.

    \item For binary and multi-class problems, we show that accurate modeling of aleatoric and epistemic uncertainties with state-of-the-art methods and combining them helps to improve the quality of selective prediction. In particular, we propose a new hybrid uncertainty quantification method, HUQ-2, that combines aleatoric and epistemic uncertainty and further boosts the quality.

    \item For the multi-label MCP task, we observe a significant increase in prediction accuracy from selective prediction on the label level compared to the instance level. 
  \end{enumerate}

  We organize the rest of the paper as follows. In Section~\ref{sec:results}, %Results,
  we briefly describe the experimental setup for evaluating UQ methods and present the results of selective prediction with multiple uncertainty quantification approaches for medical diagnosis tasks. Discussion of Section~\ref{sec:discussion} %Discussion,
  addresses challenges such as the limited application of UQ methods to medical text-based tasks and the limited use of label-level uncertainty in multi-label classification. We explore how label-wise and instance-wise selective prediction can enhance diagnostic reliability while reflecting on the study's limitations and contributions. Finally, in Section~\ref{sec:methods}, %Methods,
  we review related work on UQ, describe the baseline and advanced methods used for analysis, and provide a comprehensive overview of the experimental setup, including datasets, tasks, metrics, and implementation details.

\section{Results}
\label{sec:results}
  We conducted experiments on several medical tasks using diverse datasets to evaluate uncertainty quantification methods in selective prediction. These tasks include binary mortality prediction derived from MIMIC-III~\cite{Johnson2016-tg} dataset, multi-label medical code prediction (MCP) from MIMIC-IV~\cite{johnson2023mimic,mimic-4-medical} dataset focusing on the top 50 ICD-10 codes, and multi-class diagnosis classification using the collected private Outpatient Visits (OV) dataset comprising anonymized outpatient visits records from a large European city's Central Medical Information and Analysis System. Using the OV dataset in addition to public benchmarks reflects real-world scenarios, capturing the complexity and heterogeneity of clinical data, making it important to evaluate the methods under realistic conditions. Additionally, we explored mental health datasets targeting depression and anxiety detection, leveraging text-based tasks from sources like Depression-Essays (DE; \cite{stankevich2019predicting}), Depression-Social Media (DSM; \cite{ignatiev2022predicting}), data derived from the RusNeuroPsych~\cite{litvinova2018rusneuropsych} corpus (Anxiety-Letter (AL) and Anxiety-Description (AD) datasets) and the Anxiety-COVID dataset (AC; \cite{medvedeva2021lexical}). Standard preprocessing pipelines ensured consistency across datasets, facilitating fair comparisons~\cite{wen-etal-2020-medal, kuzmin2024mentaldisordersdetectionera}.

  \begin{table*}[t!] \centering%\resizebox{0.9\textwidth}{!}{
\begin{tabular}{l|c|c|c}
\toprule
\textbf{UQ Method} & \textbf{MIMIC Mortality} & \textbf{OV Dataset MCP} & \textbf{MIMIC MCP} \\
\midrule
SR & 0.38$\pm$0.01 & 0.4 & 0.25$\pm$0.01 \\
Entropy & -0.27$\pm$0.01 & 0.37 & 0.17$\pm$0.01 \\
Delta & -0.27$\pm$0.01 & 0.38 & 0.14$\pm$0.01 \\
Beta & 0.37$\pm$0.02 & 0.4 & -0.07$\pm$0.00 \\\midrule
MC (PV) & \textbf{0.49$\pm$0.01} & 0.28 & -0.04$\pm$0.01 \\
MC (SMP) & \underline{0.48$\pm$0.01} & \underline{0.44} & -0.23$\pm$0.01 \\
MC (BALD) & \textbf{0.49$\pm$0.01} & 0.33 & -0.12$\pm$0.02 \\\midrule
NUQ ep. & 0.44$\pm$0.03 & 0.05 & 0.14$\pm$0.00 \\
DDU & 0.39$\pm$0.02 & 0.0 & 0.24$\pm$0.01 \\
RDE & 0.45$\pm$0.02 & 0.17 & 0.15$\pm$0.00 \\
MD & 0.37$\pm$0.02 & -0.06 & \underline{0.26$\pm$0.01} \\\midrule
HUQ-DDU & 0.45$\pm$0.01 & \underline{0.41} & \underline{0.30$\pm$0.01} \\
HUQ2-DDU & \underline{0.47$\pm$0.01} & 0.4 & \textbf{0.34$\pm$0.01} \\
HUQ-RDE & 0.45$\pm$0.01 & \textbf{0.47} & \underline{0.26$\pm$0.01} \\
HUQ2-RDE & 0.45$\pm$0.02 & \underline{0.44} & 0.24$\pm$0.05 \\
HUQ-MD & 0.45$\pm$0.01 & 0.4 & \underline{0.31$\pm$0.01} \\
HUQ2-MD & \underline{0.47$\pm$0.01} & 0.4 & \textbf{0.34$\pm$0.01} \\
\bottomrule
\end{tabular}
%}
\caption{\label{tab:mimic_50} Results for the selective classification task for MIMIC mortality, OV, and MIMIC MCP datasets. We use normalized RC-AUC$\uparrow$ on the first 50\% of the curve metric for the MIMIC mortality detection and OV datasets. For the MIMIC medical code prediction task, we use the area under the first 50\% of the F1-micro rejection curve (FR-AUC$\uparrow$). The best results for each dataset are shown in bold. We underline top-3 methods after the best.}
\end{table*}
  
  We fine-tuned state-of-the-art Transformer-based architectures~\cite{vaswani2023attentionneed} such as GatorTron-base~\cite{yang2022large}, Clinical-Longformer\cite{clinical-longformer}, and private Longformer~\cite{longformer} for model training tailored to specific tasks. At the same time, we employed RuBioRoBERTa~\cite{yalunin2022rubioroberta} for mental health tasks. These models served as the foundation for computing uncertainty metrics, emphasizing their role in evaluating selective prediction. We used RC-AUC~\cite{rc-auc} for binary and multi-class tasks and introduced the FR-AUC metric for multi-label tasks. This novel metric measures the area under the F1-micro rejection curve and evaluates the model's ability to reject predictions at the label level based on uncertainty scores. To enhance interpretability, we normalized RC-AUC and FR-AUC against random and oracle uncertainty scores, focusing on rejection scenarios with up to 50\% coverage. This approach underscores the models' effectiveness in balancing predictive performance with robust uncertainty management. Importantly, any values of normalizedRC-AUC and FR-AUC greater than zero indicate the benefit of the uncertainty-informed selective prediction over the random abstention.

\subsection{General results for Instance-wise Selective Prediction}
\subsubsection{Medical Tasks}
  Table~\ref{tab:mimic_50} presents the results for the first 50\% of the curves for the selective prediction task for the MIMIC mortality, OV, and MIMIC MCP datasets. Table~\ref{tab:mimic_100} of Appendix~\ref{appendix:results} presents the results on the full curves. For the mortality prediction task, the HUQ methods significantly improved over the baselines on the entire rejection curve, outperforming the MC dropout at 2\% and the RDE at 4\%. However, in the initial 50\% of the rejected instances, MC Dropout is slightly superior to HUQ. The results for the OV dataset demonstrate that the HUQ-RDE method outperforms MC (SMP) by 3\% and SR by 7\% as evaluated by RC-AUC 50\%, showing the best result in a real-world scenario. Finally, the MIMIC dataset's medical code prediction task results demonstrate that HUQ2-DDU outperforms the best-performing baseline method MD by 8\%. Similarly, the HUQ2-DDU method significantly outperforms other methods on the full curve. Furthermore, MC Dropout with various aggregation techniques performs worse than a random choice.
  \begin{table*}[t!]
\centering%\resizebox{0.9\textwidth}{!}{
\begin{tabular}{l|c|c|c|c|c}
\toprule
\textbf{UQ Method} & \textbf{DE} & \textbf{DSM} & \textbf{AL} & \textbf{AD} & \textbf{AC} \\
\midrule

SR & 0.53$\pm$0.13 & 0.01$\pm$0.19 & 0.02$\pm$0.11 & -0.29$\pm$0.15 & 0.16$\pm$0.16 \\
Entropy & 0.53$\pm$0.13 & 0.01$\pm$0.19 & 0.02$\pm$0.11 & -0.29$\pm$0.15 & 0.16$\pm$0.16 \\
Delta & 0.53$\pm$0.13 & 0.01$\pm$0.19 & 0.02$\pm$0.11 & -0.29$\pm$0.15 & \underline{0.16$\pm$0.16} \\
Beta & 0.07$\pm$0.42 & 0.04$\pm$0.18 & 0.02$\pm$0.13 & -0.08$\pm$0.26 & 0.01$\pm$0.32 \\ \midrule
MC (PV) & 0.37$\pm$0.27 & 0.01$\pm$0.21 & 0.02$\pm$0.34 & -0.16$\pm$0.27 & 0.02$\pm$0.13 \\
MC (SMP) & 0.43$\pm$0.28 & 0.08$\pm$0.13 & 0.00$\pm$0.11 & -0.28$\pm$0.19 & \underline{0.17$\pm$0.19} \\
MC (BALD) & 0.34$\pm$0.29 & 0.01$\pm$0.20 & 0.02$\pm$0.37 & -0.15$\pm$0.27 & 0.02$\pm$0.12 \\ \midrule
NUQ ep. & 0.45$\pm$0.07 & 0.09$\pm$0.18 & -0.01$\pm$0.25 & \underline{-0.06$\pm$0.24} & 0.16$\pm$0.12 \\
DDU & 0.53$\pm$0.08 & \underline{0.17$\pm$0.18} & -0.02$\pm$0.20 & \underline{-0.04$\pm$0.25} & \underline{0.17$\pm$0.10} \\
RDE & \underline{0.55$\pm$0.08} & -0.06$\pm$0.14 & -0.11$\pm$0.21 & -0.06$\pm$0.29 & 0.12$\pm$0.10 \\
MD & 0.10$\pm$0.11 & 0.03$\pm$0.10 & \textbf{0.11$\pm$0.16} & \textbf{0.11$\pm$0.17} & 0.09$\pm$0.14 \\ \midrule
HUQ-DDU & \underline{0.55$\pm$0.06} & \underline{0.13$\pm$0.17} & 0.01$\pm$0.12 & -0.14$\pm$0.21 & \textbf{0.18$\pm$0.10} \\
HUQ2-DDU & 0.54$\pm$0.06 & \textbf{0.24$\pm$0.20} & \underline{0.06$\pm$0.09} & -0.15$\pm$0.25 & 0.10$\pm$0.09 \\
HUQ-RDE & \textbf{0.57$\pm$0.07} & 0.01$\pm$0.19 & 0.02$\pm$0.11 & -0.20$\pm$0.20 & 0.14$\pm$0.10 \\
HUQ2-RDE & \underline{0.55$\pm$0.06} & \underline{0.11$\pm$0.24} & \underline{0.04$\pm$0.08} & \underline{-0.01$\pm$0.28} & 0.05$\pm$0.08 \\
HUQ-MD & 0.49$\pm$0.15 & -0.04$\pm$0.22 & 0.01$\pm$0.10 & -0.17$\pm$0.23 & 0.11$\pm$0.14 \\
HUQ2-MD & 0.50$\pm$0.15 & -0.03$\pm$0.16 & \underline{0.04$\pm$0.17} & -0.16$\pm$0.28 & 0.05$\pm$0.22 \\
\bottomrule
\end{tabular}
%}
\caption{\label{tab:psy_ue_rubioroberta_50}Results for the selective classification task for mental disorder detection datasets. The best results for each dataset are shown in bold. The top-3 methods after the best are underlined. The metric is normalized RC-AUC$\uparrow$ on the first 50\% of the curve.}
\end{table*}

\begin{figure*}[t]
    \footnotesize
    \centering
    \begin{minipage}[h]{0.45\linewidth}
    \center{\includegraphics[width=5.5cm]{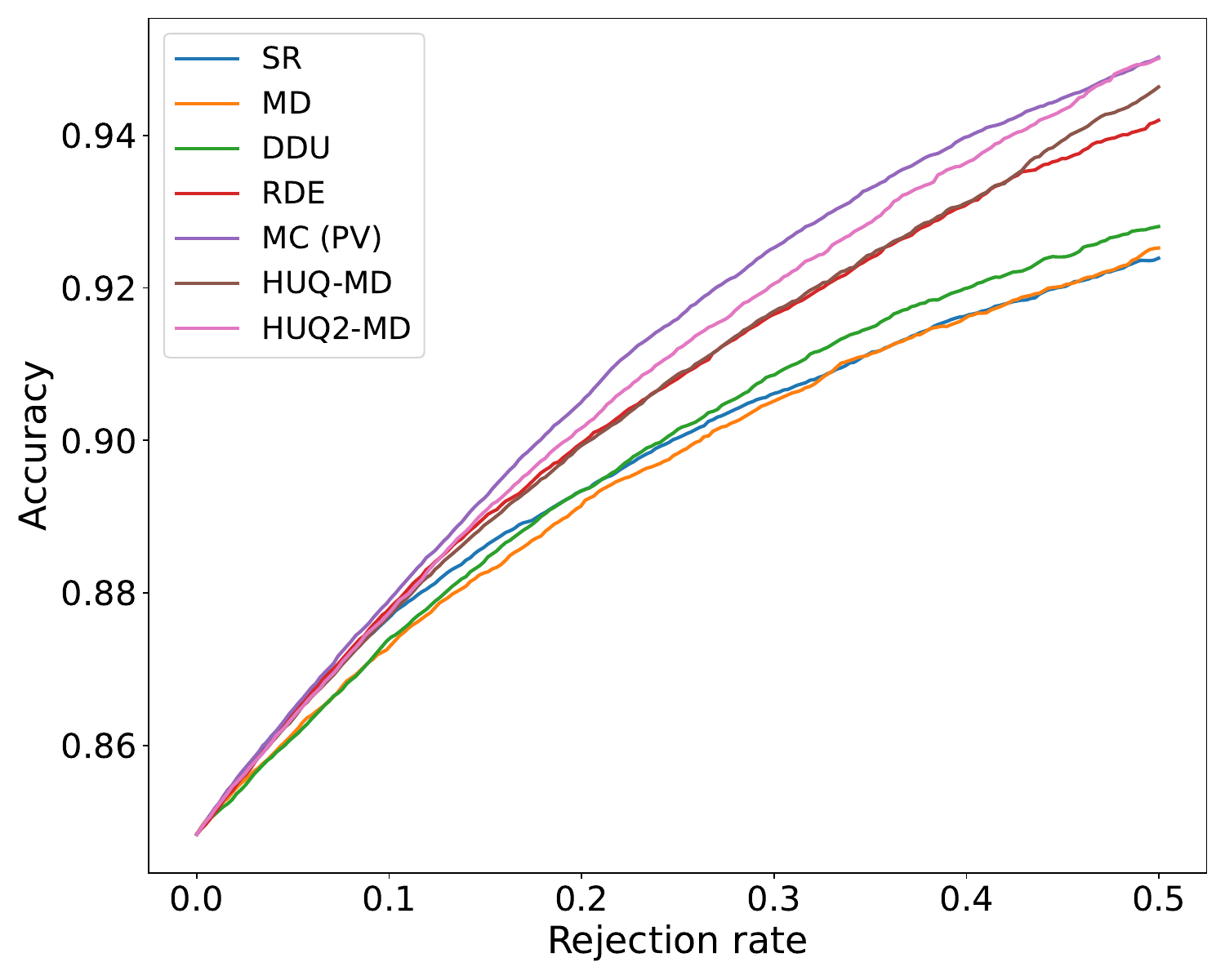} \\
    a) MIMIC-III mortality prediction 
    }
    \end{minipage}
    \begin{minipage}[h]{0.45\linewidth}
    \center{\includegraphics[width=5.5cm]{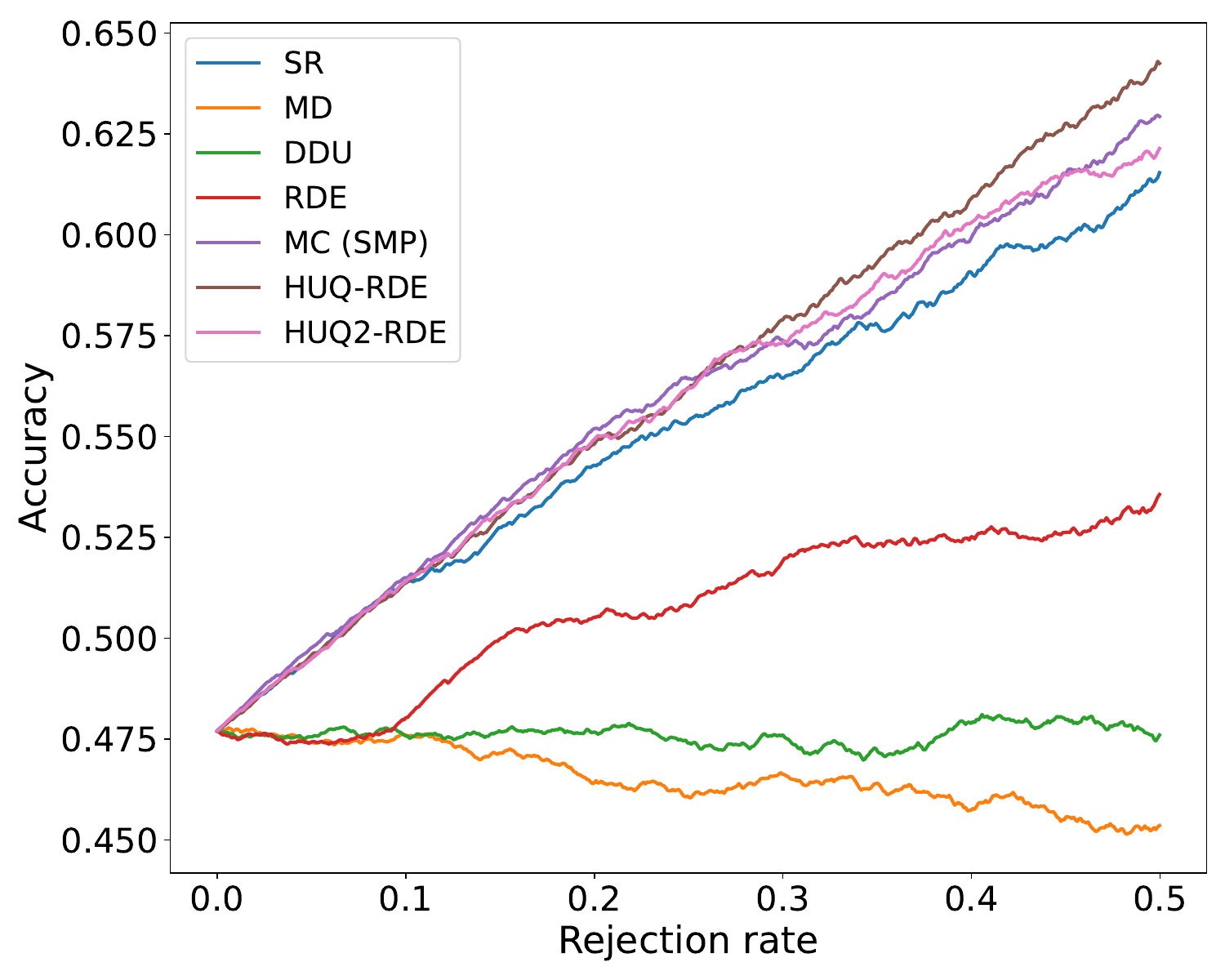} \\
    b) OV medical code prediction 
    }
    \end{minipage}
    \begin{minipage}[h]{0.45\linewidth}
    \center{\includegraphics[width=6.cm]{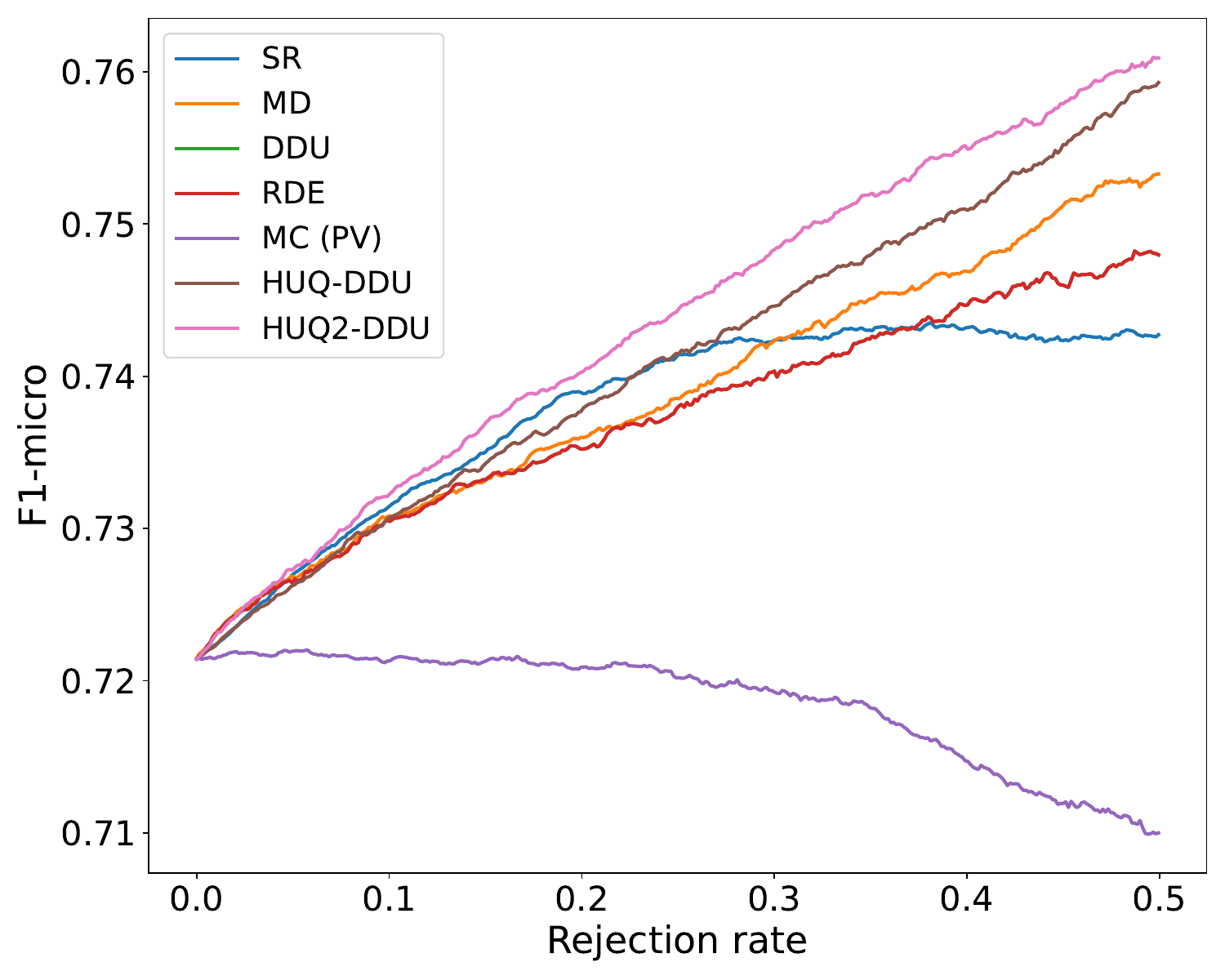} \\
    c) MIMIC-IV medical code prediction
    }
    \end{minipage}
    \caption{\label{fig:rejection_curves}Rejection curves for the selected methods for the considered tasks.}
  \end{figure*}

  Figure~\ref{fig:rejection_curves} presents the first 50\% of the rejection curves for the considered task for the selected methods. Figure~\ref{fig:full_rejection_curves} of Appendix~\ref{appendix:results} presents the full rejection curves. We choose the best methods from each group based on their performance. For the mortality prediction task, the curves illustrate that MC (PV) in the first part of the curve is marginally higher than HUQ2-MD. Nevertheless, these methods are significantly higher than SR and MD and slightly higher than RDE. 

  For the OV dataset, we see that the MD and DDU methods cannot enhance the accuracy and perform only marginally better than random estimates. Furthermore, we see that HUQ-RDE on the initial part of the curve performs comparably with MC (SMP). Nevertheless, starting from 30\% of rejected instances, HUQ-RDE shows notable improvement over all other methods. Moreover, HUQ-RDE enhances the overall accuracy of the model by 15\% when rejecting 40\% of the most uncertain instances. These improvements are crucial for safety-critical applications in the medical field, where methods must be tested on real-world data to ensure reliability and effectiveness in complex and diverse clinical scenarios.
  
  For the medical code prediction task, the MC (PV) curve is considerably behind the random estimates and, therefore, is unsuitable for this particular task. Moreover, the curve for the HUQ2-DDU is markedly higher than that of all other selected methods. Overall, the results demonstrate the robustness of the HUQ methods, which are the best or second-best methods for all datasets, while introducing only a small amount of overhead compared to the MC Dropout.

\subsubsection{Mental Disorders Detection Task}
  Table~\ref{tab:psy_ue_rubioroberta_50} presents the results for the first 50\% of the curves for the selective prediction task on the datasets for mental disorders detection. Table~\ref{tab:psy_ue_rubioroberta_100} of Appendix~\ref{appendix:results} presents the results on the full curve. The best results were achieved on the DE dataset with HUQ2-RDE, which improves over the SR baseline by 20\% by RC-AUC. The initial 50\% HUQ2-RDE is comparable with RDE and still outperforms all other methods. HUQ2-DDU achieves the best results in DSM on both the full rejection curve and the initial part, while HUQ-DDU performs better than all other methods in the AC data set. On the contrary, hybrid methods on the AL and AD datasets perform worse than density-based methods. MD achieved the best results on these datasets on both the initial 50\% and 100\% of the rejection curve. Overall, hybrid methods and density-based methods show the best performance over all datasets for mental disorders detection. At the same time, MC Dropout only slightly improves over baseline on DSM and AD.

  \begin{figure*}[t]
    \footnotesize
    \centering
    \begin{minipage}[h]{0.45\linewidth}
    \center{\includegraphics[width=5.5cm]{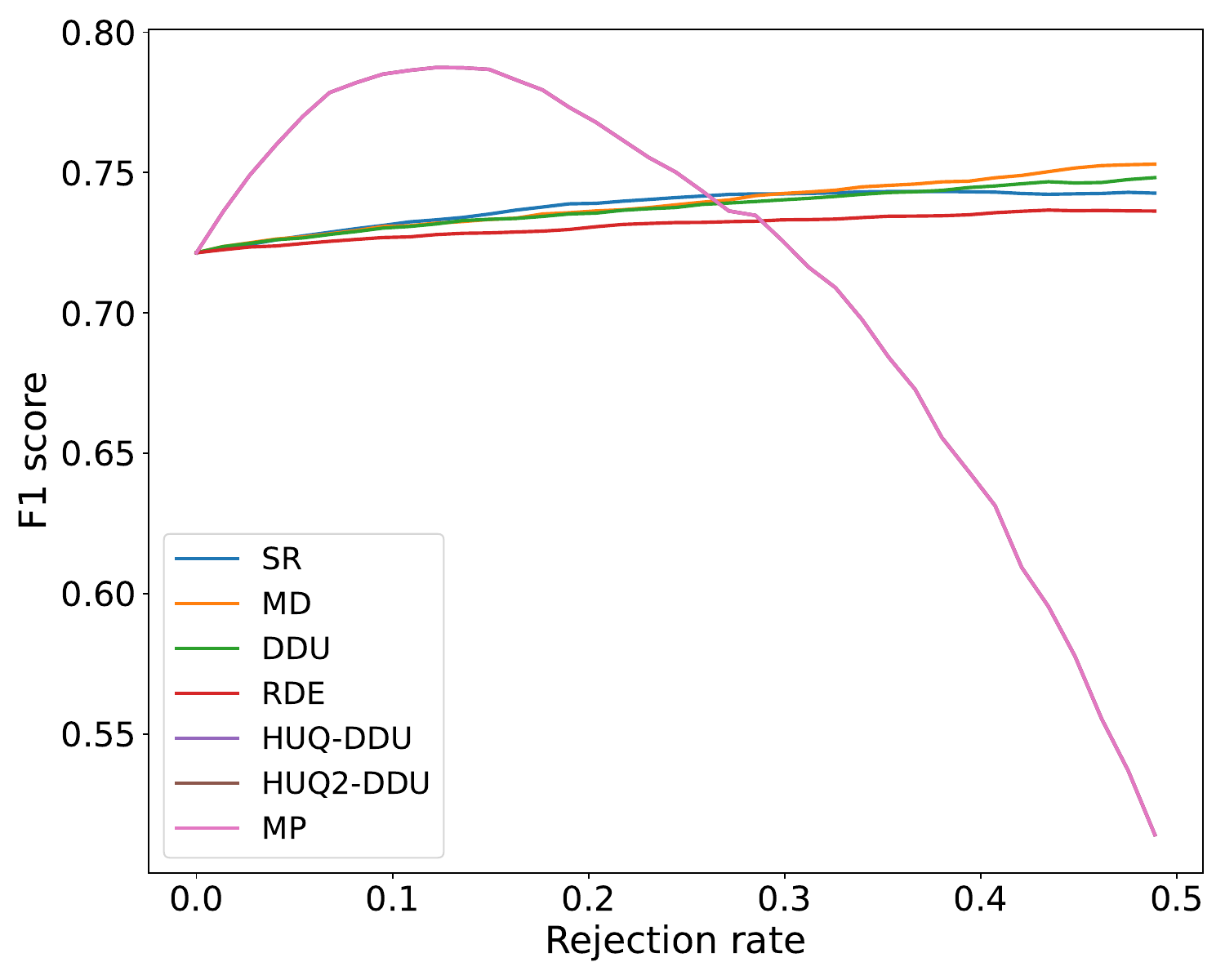} \\
    % a) F1 score 
    }
    \end{minipage}
    \begin{minipage}[h]{0.45\linewidth}
    \center{\includegraphics[width=5.5cm]{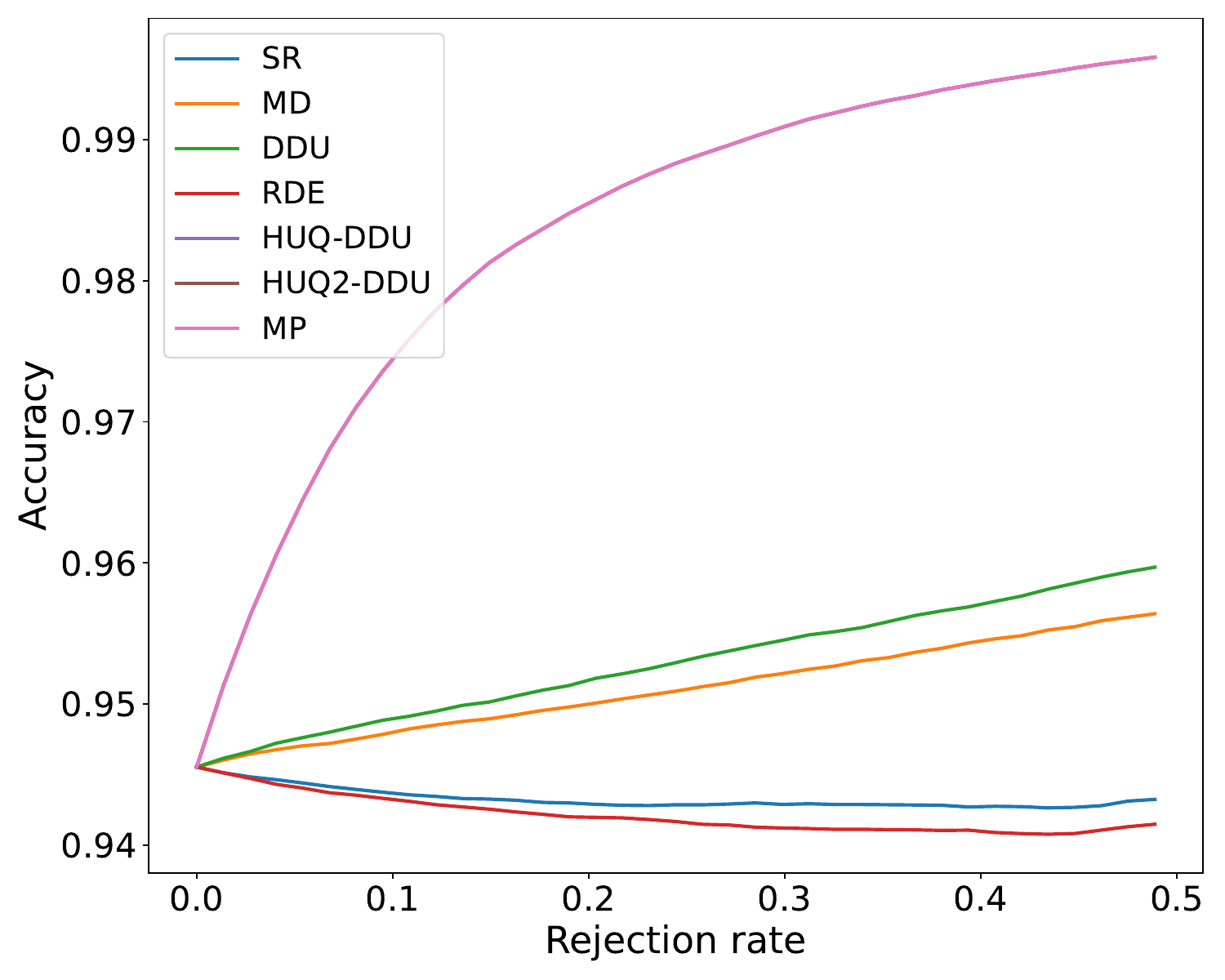} \\
    % b) Accuracy
    }
    \end{minipage}
    \caption{\label{fig:label_rejection_curves} Rejection curves for the selected methods for the MIMIC medical code prediction task for general rejection methods vs label-wise MP approach. HUQ hyperparameters are fitted using the accuracy rejection metric. The HUQ and HUQ-2 methods overlap with the MP method due to the selected hyperparameters on the validation set.}
  \end{figure*}

\subsection{Results for Label-wise Selective Prediction for Multi-label Data}
  We summarize the results for label-wise selective prediction for multi-label data in Figure~\ref{fig:label_rejection_curves}. The patterns observed here are interesting and differ substantially from the ones for instance-wise selective prediction. In Figure~\ref{fig:label_rejection_curves}-(a), we observe that in terms of accuracy, the label-wise approach gives a consistent and very significant boost compared to label-wise methods. For the F1-micro score, the situation is more involved; see Figure~\ref{fig:label_rejection_curves}-(b). On the one hand, one can improve significantly over instance-wise methods by rejecting just 10\% of points. At the same time, we observe a notorious drop in the F1-micro score for higher rejection rates. The reason is that individual class probabilities are not well calibrated, and using common thresholds for them leads to throwing out too many true positive examples for some of the classes. The HUQ and HUQ-2 methods overlap with the MP method due to the hyperparameters selected on the validation set.

\section{Discussion}
\label{sec:discussion}
  % In this work we demonstrate the capabilities of state-of-the-art uncertainty quantification methods in the task of selective prediction (or abstention) for medical diagnosis based on the medical texts. We show that abstention allows to significantly boost the prediction quality in the wide variety of tasks. Importantly, to achieve this results, it is crucial to correctly model various types of uncertainty and carefully consider the peculiarities of the particular prediction problem.

  % Based on the performed research, we see a great potential in employing selective prediction approaches in the real-world automated medical diagnosis pipelines. It will allow to increase the quality of the diagnostics by referring complex uncertain cases to the doctor. At the same time, it will allow to greatly boost confidence in such systems as the patients will know that the diagnosis won't be based solely on the automated system. 

  % For the future research, we see great importance in further studies on uncertainty quantification for the text data. While ultimate uncertainty measure might not exist, the results might be improved by considering parameterized UQ measures with parameters being optimized for the particular applied problem. At the same time, label-wise approaches seem to be very relevant for applications and also very promising for further studies.

  In deploying artificial intelligence systems in clinical settings, the ability to estimate and interpret uncertainty is crucial to ensure reliable and trustworthy decision-making. Despite significant advances in deep learning, the inherent variability of medical data and the high stakes of clinical decisions require robust uncertainty estimation methods. Techniques such as Bayesian approaches, Monte Carlo dropout, and ensemble methods have shown promise in quantifying uncertainty. However, challenges remain with comparing different uncertainty quantification methods across various classification tasks involving medical texts. These challenges arise from the diversity of tasks, such as disease classification and ICD code prediction, as well as differences in how methods handle ambiguity, domain-specific terminology, and the variability of annotations. Addressing these challenges is vital to bridge the gap between technical developments and clinical adoption, ultimately providing physicians with more reliable tools to support decision-making and improve patient communication. To this end, we demonstrate the capabilities of state-of-the-art uncertainty quantification methods in selective prediction (or abstention) for medical diagnosis based on the medical texts. We show that abstention allows us to boost the prediction quality significantly in various tasks. To achieve these results, it is crucial to correctly model various types of uncertainty and carefully consider the peculiarities of the particular prediction problem.

  % Uncertainty quantification methods, such as softmax response, Monte Carlo dropout, and density-based approaches, have been developed to improve model reliability by estimating prediction confidence. Combining multiple strategies, hybrid methods have further advanced UQ by capturing model and data uncertainty. While these approaches have shown promise in tasks like in-hospital mortality prediction and ICD code assignment, their application to medical texts remains limited. Challenges persist due to the variability of medical language, the complexity of clinical documentation, and the need for robust evaluation across diverse tasks.

  However, significant obstacles remain in applying UQ to medical text classification tasks, which we address in this work. First, the application of UQ methods to medical text-based tasks has been limited, reducing their demonstrated utility. To overcome this, we systematically evaluate state-of-the-art UQ techniques across various tasks, including mortality prediction, mental disorder detection, and multi-class and multi-label medical code classification.

  Second, existing approaches often fail to effectively model and combine aleatoric (data-related) and epistemic (model-related) uncertainties, which reduces their predictive performance. We address this by introducing HUQ-2, a novel hybrid UQ method that combines these uncertainties, significantly improving selective prediction quality for binary and multi-class classification tasks.

  Third, current UQ techniques in multi-label classification are typically applied at the instance level, missing the opportunity to leverage label-level uncertainty. We demonstrate that performing selective prediction at the label level substantially enhances accuracy for multi-label tasks, such as medical code prediction.

  Finally, prior work often focuses on narrow, task-specific evaluations, limiting UQ methods' generalizability. In response, we evaluate UQ techniques on various medical classification tasks, showcasing their robustness and effectiveness across binary, multi-class, and multi-label problems. We use public benchmarks and a new real-world dataset to ensure practical applicability and relevance.

  Despite the encouraging results obtained with the hybrid uncertainty quantification methods, it is important to acknowledge their limitations. Firstly, none of the considered methods can guarantee high uncertainty for erroneous predictions. Even with the most advanced uncertainty quantification techniques, there remains a possibility of estimating a high uncertainty for a correct prediction and vice versa. Secondly, hybrid uncertainty quantification methods require the utilization of a validation dataset for hyperparameter tuning. This reliance on a validation dataset presents potential limitations in scenarios where labeled data is unavailable. Lastly, our research explores uncertainty quantification methods for the classification tasks with the encoder-only models and does not address the large language models (LLMs). However, modern LLMs have shown remarkable success in solving specific applications within the medical domain~\cite{medqallms}, such as medical question answering.

  In conclusion, we see great potential in employing selective prediction approaches in real-world automated medical diagnosis pipelines, allowing doctors to refer complex, uncertain cases to the doctor and increasing the quality of the diagnostics. At the same time, it will significantly boost confidence in such systems, as patients will know that the diagnosis will not be based solely on the automated system.
  
  For future research, we see great importance in further studies on uncertainty quantification for the text data. Although an ultimate uncertainty measure might not exist, the results might be improved by considering parameterized UQ measures with parameters optimized for the particular applied problem. At the same time, label-wise approaches are very relevant for applications and promising for further studies. One more direction for future work is to explore uncertainty quantification methods for the LLMs in the medical domain.

\section{Methods}
\label{sec:methods}

\subsection{Related Work}
  % Works examining uncertainty quantification in medicine can be divided into two groups. The first group of articles examines the we call \textit{post-facto} uncertainty. In these works, the authors calculate the degree of model uncertainty after each prediction, and the resulting value is analyzed. In the medical field, these approaches are also used to assess uncertainty. Thus, in the work~\cite{10.1145/3368555.3384457}, the authors, using ensembles of models~\cite{lakshminarayanan2017simple} and various Bayesian neural networks~\cite{DBLP:journals/corr/FortunatoBV17}, studied the uncertainty of models in the tasks of in-patient mortality prediction and diagnosis prediction at discharge. The findings demonstrated that significant diversity in predictions and decisions tailored to individual patients may signal the potential fragility of a model's decision. This type of decision presents a chance to recognize supplementary data, aiming to mitigate the extent of uncertainty associated with the model.

  Uncertainty quantification methods have evolved to include various techniques to improve model reliability. For example, softmax response~\cite{selective-classification}, one of the simplest UQ methods, estimates the confidence of a model by the softmax output, assuming high probability indicates high certainty. Additionally, a group of Monte Carlo Dropout approaches~\cite{pmlr-v48-gal16, kampffmeyer2016semantic, gal2017deep} addresses this problem by applying dropout during inference, generating multiple predictions from the same model and estimating uncertainty from their variability. Another group of methods, density-based methods~\cite{lee2018simple, yoo-etal-2022-detection, Mukhoti2022ddu, Kotelevskii2022nuq}, focuses on analyzing the distribution of points in a model feature space using the Gaussian distribution. Moreover, a recent state-of-the-art method named Hybrid Uncertainty Quantification (HUQ; \cite{vazhentsev-etal-2023-hybrid}) integrates different strategies, such as combining density-based methods with the neural softmax response, to capture both model and data uncertainties comprehensively.

  While the described methods are well-established in uncertainty quantification, their application in medicine, particularly in electronic health records (EHRs), remains limited. However, there is a growing trend toward using UQ to enhance decision-making processes. For example, the work~\cite{10.1145/3368555.3384457} explored the uncertainty of models in the tasks of in-patient mortality prediction and diagnosis prediction at discharge. This paper considered using ensembles of models~\cite{lakshminarayanan2017simple} and various Bayesian neural networks~\cite{DBLP:journals/corr/FortunatoBV17}.   After each prediction, they calculated the degree of model uncertainty and analyzed the resulting value. The findings demonstrated that significant diversity in predictions and decisions tailored to individual patients may signal the potential fragility of a model's decision. This type of decision presents a chance to use supplementary data to mitigate the extent of uncertainty associated with the model.
  
  However, it is important not only to estimate the model's uncertainty for subsequent analysis and interpretation but also to influence the model's decision in real time. For example, in dynamic environments like emergency rooms or intensive care units, real-time inference ensures that decisions are based on current data, improving patient safety and treatment effectiveness. To solve this problem, a special approach mentioned earlier, namely \textit{selective prediction}, is used. This method relies on the model deciding to classify a sample based on confidence in its prediction. Namely, in~\cite{heo2018uncertaintyaware}, the authors used stochastic attention to obtain the ``I don't know'' prediction to solve several problems, such as mortality prediction, cardiac condition, and surgery recovery on physiological signals and vital patient information. Making this decision reduced the number of false positives/negatives and demonstrated the feasibility of using UQ for selective prediction in the medical field.
  
  In~\cite{qiu2019modeling}, when solving the problem of in-hospital mortality using the Bayesian deep learning approach on patient monitor records, the authors demonstrated that when dealing with patients who present lower uncertainty in their health conditions, the final performance metric significantly improves compared to the outcomes observed in patients with higher uncertainty. This result suggests that the model is more effective and accurate when the uncertainty value is small. These findings highlight the impact of patient uncertainty on model performance and underscore the critical role of selective prediction in the medical domain, where careful decision-making can lead to more reliable results and better overall patient outcomes.

  The authors of~\cite{li2020deep} proposed an algorithm leveraging deep kernel learning~\cite{pmlr-v51-wilson16} and Bayesian neural networks to tackle the task of detecting heart failure, diabetes, and depression using medical codes and medication information sequences. Their architecture demonstrated improved final metrics by discarding samples with high uncertainty. Moreover, it revealed the greater significance of uncertainty in smaller classes when addressing data imbalances.

  In the case of selective prediction, the prediction outcome can not only be filtered based on uncertainty assessment but also subjected to an additional classification process. For example, in~\cite{ASHFAQ2023104019}, the authors estimated model uncertainty in diagnostic predictions by demographics and clinical details using the evidential deep learning approach~\cite{sensoy2018evidential}. When the confidence level of prediction is low, the model triggers a secondary classification step by searching for the nearest neighbors within the ICD code space. This approach ensures that cases with higher uncertainty should be handled more cautiously, enhancing the overall reliability of the prediction system.

  The described works demonstrate the importance and relevance of studying the uncertainty quantification when working with selective prediction on such types of EHR data as patient monitor records~\cite{heo2018uncertaintyaware, qiu2019modeling}, visit history~\cite{ASHFAQ2023104019} and ICD code sequences~\cite{li2020deep}. However, selective prediction studies specifically on medical texts are also needed since this approach has great potential to work and obtain high-quality results in the medical field. Textual data contains context, details, and nuances that are difficult to express through codes and numeric parameters only. This information provides a more complete picture of the patient's condition, including symptoms and anamnesis description that may not be reflected in described EHR data. Integrating text information improves the accuracy of diagnostic models and prognoses, facilitating a more personalized approach to medical decisions.

  Thus, in~\cite{PELUSO2024104576}, the authors applied various selective prediction techniques to enhance the efficiency of cancer registries by automating the extraction of disease-related information from electronic pathology reports at the stages of diagnosis and surgery. They explored multiple approaches to evaluate uncertainty, including fixed confidence score, delta difference score, entropy ratio confidence score, and Bayes beta confidence score. The findings showed that these selective prediction methods successfully achieved the desired level of accuracy in a trade-off analysis, which aimed to reduce the rejection rate.

  % The studies described above demonstrate the significance and relevance of uncertainty quantification in selective prediction tasks within EHR data, especially in the context of medical texts. These works highlight the crucial role of uncertainty quantification in improving classification performance when dealing with medical data. However, there is a growing need to apply state-of-the-art UQ approaches that extend to broader important medical tasks, such as ICD code assignment or mortality prediction based on text containing medical information. These applications pose significant challenges due to the inherent uncertainty arising from the vast number of codes and the variability of medical language used in clinical documentation.

  % Our work addresses this gap by systematically comparing existing state-of-the-art uncertainty quantification methods for medical texts using public benchmarks and proprietary outpatient visits dataset. We aim to assess the effectiveness of these methods in handling the complexities of medical text classification, including tasks like ICD code assignment and mortality predictions.

\subsection{UQ Methods}
  In this section, we describe the baselines and consider various state-of-the-art uncertainty quantification methods used in our experiments, which leverage different information from the model. Specifically, we utilize Monte Carlo Dropout, density-based methods, and a hybrid uncertainty quantification approach. 

\subsubsection{Baseline Methods}
\paragraph{Softmax Response.} 
  The softmax response is the most well-known and straightforward method (SR; \cite{selective-classification}). This method uses the probability derived from the output softmax layer from the model to quantify uncertainty. SR computes the maximum probability $p(y = c \mid \xv)$ over classes $c \in C$. Lower probabilities correspond to more uncertain predictions:
  \begin{equation}
    U_{\text{SR}}(\xv) = 1 - \max_{c \in C} p(y = c \mid \xv).
  \end{equation}
  SR is fast and computationally cheap as it requires almost no additional computations. However, modern models are often overconfident in their predictions. Consequently, SR is not always an optimal choice and may not be robust across different tasks.

  For the label-wise rejection, we use the SR method for the binary classification task. For each class, we compute the probability using the sigmoid function. The uncertainty for the class $i$ of instance $\xv$ based on the maximum probability (MP) is defined as follows: 
  \begin{equation}
    U_{\text{MP}}(\xv, i) = 1 - \max{(p_i, 1 - p_i)}.
  \end{equation}

\paragraph{Delta.} 
  In cases where the model is confident, the highest probability over classes should be significantly larger than the other probabilities. Consequently, this method calculates the difference between the two highest predicted probabilities~\cite{delta}:
  \begin{equation}
    U_{\text{SR}}(\xv) = p_{(1)}(\xv) - p_{(2)}(\xv),
  \end{equation}
  where $\bar{c} = \arg\max_{c \in C} p(y = c \mid \xv)$, $p_{(1)}(\xv) = \max_{c \in C} p(y = c \mid \xv) = p(y = \bar{c} \mid \xv)$ and $p_{(2)}(\xv) = \max_{c \in C \setminus \bar{c}} p(y = c \mid \xv)$. The lower this difference is, the more uncertain the model is.
  
\paragraph{Entropy.} 
  This method considers variations in the probabilities of all classes and computes the entropy of the probability distribution derived from the output softmax layer to quantify uncertainty~\cite{gal2017deep}. Smaller entropy indicates more uncertain prediction:
  \begin{equation}
    U_{\text{Entropy}}(\xv) = -\sum_{c \in C}{p(y = c \mid \xv) \log p(y = c \mid \xv)}.
  \end{equation}
  
\paragraph{Bayes Beta.} 
  This method utilizes the Bayes theorem for the binary variable, where correct and incorrect predictions are mutually exclusive occurrences. The uncertainty score, according to the Bayes Beta method~\cite{PELUSO2024104576}, is as follows:
  \begin{equation}
    U_{\text{Beta}}(\xv) = 1 - \frac{p(y\mid\text{correct}) \, p(\text{correct})}{p(y\mid\text{correct}) \, p(\text{correct}) + p(y\mid\text{incorrect}) \, p(\text{incorrect})},
  \end{equation}
  where $y = \arg\max_{c \in C}{p(y = c \mid \xv)}$ is a prediction for a test instance $\xv$, $p(\text{correct})$ and $p(\text{incorrect})$ are prior probabilities and are estimated by the frequencies of the total number of correct or incorrect predictions on the validation data, $p(y\mid\text{correct})$ and $p(y\mid\text{incorrect})$ are estimated assuming $p(y\mid\text{correct})\sim Beta(\alpha_{c}, \gamma_{c})$ and $p(y\mid\text{incorrect})\sim Beta(\alpha_{inc}, \gamma_{inc})$ and the hyperparameters $\alpha_{c}, \gamma_{c}, \alpha_{inc}, \gamma_{inc}$ are obtained via maximum likelihood estimation on the validation data.

\subsubsection{Monte Carlo Dropout}
  This group of methods is based on multiple stochastic inference with the activated dropout in all hidden layers of the model. Supposing we perform $T$ stochastic forward passes, the following approaches are used to estimate uncertainty based on the multiple predictions via MC dropout:

\noindent\textbf{Sampled maximum probability} (SMP) is defined as follows:
  \begin{equation}
    U_{\text{SMP}}(\xv) = 1 -  \max_{c \in C} \frac{1}{T}\sum_{t=1}^T  p_t^{c}(\xv) ,
  \end{equation}
  where $p^{c}_t(\xv)$ represents the probability of the class $c$ for the $t$-th stochastic forward pass.

\noindent\textbf{Probability variance} (PV; \cite{kampffmeyer2016semantic,gal2017deep}) is defined as following: 
  \begin{equation}
    U_{\text{PV}}(\xv) = \frac{1}{C} \sum_{c = 1}^C \biggl(\frac{1}{T - 1} \sum_{t = 1}^T {\bigl(p^c_t(\xv) - \bar{p}^c(\xv)\bigr)^2} \biggr),
  \end{equation}
  where $\bar{p}^c(\xv) = \frac{1}{T} \sum_{t = 1}^T p^{c}_{t}(\xv)$ is the probability for a given class $c$ averaged across $T$ stochastic forward passes.

\noindent\textbf{Bayesian active learning by disagreement} (BALD; \cite{houlsby2011bayesian}) is defined as following:
  \begin{equation}
    U_{\text{BALD}}(\xv) = -\sum_{c = 1}^C \bar{p}^c(\xv) \log \bar{p}^c(\xv) + \frac{1}{T} \sum_{c = 1}^C \sum_{t = 1}^T p^{c}_{t}(\xv) \log p^{c}_{t}(\xv).
  \end{equation}
  These methods are well-established and represent strong baselines. Previously, it was shown that MC dropout outperforms the SR baseline only when all dropout layers in the Transformer model are activated~\cite{shelmanov-etal-2021-certain}. Therefore, in our work, we follow the same setting. We also note that MC dropout requires multiple stochastic forward passes, which makes it computationally expensive, incurring a considerable overhead. This requirement makes MC dropout less suitable for practical applications.

\subsubsection{Density-based Methods}
\label{sec:density_based_methods}
  Recently, various computationally cheap alternatives to MC dropout were proposed that do not require multiple inferences of the model. Among the most notable approaches are density-based methods. These methods assume the class-conditional distribution of the hidden representations of instances in the training dataset follows a multivariate Gaussian distribution.

\paragraph{Mahalanobis Distance.}
  One of the foundation methods from this group is Mahalanobis Distance (MD; \cite{lee2018simple}). For a given training dataset $\DC$, $h(\xv)$ represents a hidden representation of an instance $\xv$ from the trained model. The penultimate layer of the model is used to extract embeddings. MD fits $C$ Gaussians for each class $c\in C$ using the training dataset $\DC$. We compute class centroids $\{\mu_c\}_{c \in C}$ and shared across classes covariance matrix $\Sigma$. Finally, the Mahalanobis distance between $h(\xv_)$ and the closest Gaussian represents the uncertainty of the model prediction:
  \begin{equation}
    U_{\text{MD}}(\xv) = \min_{c \in C} (h(\xv) - \mu_{c})^{T} \Sigma^{-1} (h(\xv) - \mu_{c}).
  \end{equation}

\paragraph{Robust Density Estimation.}
  Recent work introduces {Robust Density Estimation} (RDE; \cite{yoo-etal-2022-detection}), which is a modification of MD. This method computes the covariance matrix $\Sigma_c$ for each class using the Minimum Covariance Determinant (MCD) estimation~\cite{Rousseeuw84leastmedian} and reduces the dimensionality of the hidden representations via PCA decomposition with an RBF kernel. The use of MCD aims to reduce the impact of outliers for density estimation by minimizing the determinant of the covariance matrix using a subset of the entire training dataset.

\paragraph{Deep Deterministic Uncertainty.}
  The next approach is Deep Deterministic Uncertainty (DDU; \cite{Mukhoti2022ddu}), which fits a Gaussian Mixture Model (GMM) $p(h(\xv)\mid y)$ with a single mixture component per class, which is also a modification of the MD method. In order to  quantify uncertainty, the probability density of $h(\xv)$ under the GMM is used:
  \begin{equation}
    U_{\text{DDU}}(\xv) = \sum_{c \in C} p(h(\xv) \mid y = c) \,\, p(y = c), ~~ \text{where}
  \end{equation}
  $p(h(\xv) \mid y = c) \sim \mathcal{N}(h(\xv) \mid \mu_c,\,\Sigma_c)$ and $p(y = c) = \frac{1}{|\DC|} \sum_{(\xv_i, y_i) \in \DC} \mathbf{1}[y_i = c]$.

\paragraph{Nonparametric Uncertainty Quantification.}
  The final approach in this group is Nonparametric Uncertainty Quantification (NUQ; \cite{Kotelevskii2022nuq}). This method constructs an asymptotic approximation of the expected value of an upper bound for the total Bayes risk. It applies the Nadaraya-Watson estimator with an RBF kernel to obtain conditional label probabilities $\hat{p}(y = c \mid \xv)$ and the variance $\hat{\sigma}^2_c (\xv)$, leading to the following epistemic uncertainty score: 
  \begin{equation}
    U_{\text{NUQ}}(\xv) \!=\! 2 \sqrt{\frac{2}{\pi}} \hat{\tau}(\xv),
    \quad
    \hat{\tau}^2(\xv) \!\! = \!\! \frac{\tilde{C}}{|\DC|} \frac{\max_c \hat{\sigma}_c^2(\xv)}{\hat{p}(\xv)},
  \end{equation}
  where $\hat{p}(\xv)$ is the probability density estimated with a kernel density estimator and $\tilde{C}$ is a constant of the kernel. 
  $\tilde{C} = \frac{h^d}{2\sqrt{\pi}} $ is a constant computed for an RBF kernel, and $d$ is a dimension of the hidden representation.

  Contrary to MC dropout, density-based methods do not require multiple inferences. Therefore, these methods are fast and computationally cheap. Additionally, density-based methods provide good performance for out-of-distribution detection~\cite{PodolskiyLBAP21Revisiting} and, in some cases, for selective prediction~\cite{vazhentsev-etal-2022-uncertainty}.

\subsection{Hybrid Methods}

\paragraph{Hybrid Uncertainty Quantification (HUQ).} 
  The state-of-the-art Hybrid Uncertainty Quantification (HUQ; \cite{vazhentsev-etal-2023-hybrid}) method combines epistemic uncertainty quantification, represented by density-based methods with SR, into a single score. The main idea behind HUQ is to use different types of uncertainty depending on whether the instance lies close to the out-of-distribution area of the feature space or around the discriminative border between classes. 

  Firstly, we need to define the set of in-distribution instances from $\DC$ as follows: $\DC_{\text{ID}} = \{\xv \in \DC\colon U_{\text{E}}(\xv) \leq \delta_{\min}\}$. We also defined the set of arbitrary in-distribution instances $\XSet_{\text{ID}} = \{\xv\colon U_{\text{E}}(\xv) \leq \delta_{\min}\}$ and ambiguous in-distribution instances (instances that lie around the discriminative border of the trained classifier) $\XSet_{\text{IDA}} = \{\xv {\in \XSet_{\text{ID}}}\colon U_{\text{A}}(\xv) > \delta_{\max}\}$ using $\delta_{\min}$, $\delta_{\max}$ are thresholds selected on the validation dataset. 

  Consider we compute measures of aleatoric $U_{\text{A}}(\xv)$ and epistemic $U_{\text{E}}(\xv)$ uncertainty. In order to make different uncertainty scores comparable, we define a ranking function $R(\uv, \mathfrak{D})$ as a rank of $\uv$ over a sorted dataset $\mathfrak{D}$, where $\uv_1 > \uv_2$ implies $R(\uv_1, \mathfrak{D}) > R(\uv_2, \mathfrak{D})$. For a given measure of aleatoric and epistemic uncertainty, we compute the total uncertainty $U_{\text{T}}(\xv)$ as a linear combination $U_{\text{T}}(\xv) = \!\!(1 - \alpha) R(U_{\text{E}}(\xv), \DC) + \alpha R(U_{\text{A}}(\xv), \DC)$, where $\alpha$ is a hyperparameter selected on the validation dataset. As a result, we define HUQ as follows:
  \begin{equation}
    U_{\text{HUQ}}(\xv)
    = 
    \begin{cases}
      R(U_{\text{A}}(\xv), \DC_{\text{ID}}), \forall \xv \in \XSet_{\text{ID}} \setminus \XSet_{\text{AID}}, \\
      R(U_{\text{A}}(\xv), \DC),  \forall \xv \in \XSet_{\text{AID}}, \\
      U_{\text{T}}(\xv), \forall \xv \notin \XSet_{\text{ID}}.
    \end{cases}
  \end{equation}
  Recently, it was shown that the HUQ method provides the best performance in case of ambiguous tasks~\cite{vazhentsev-etal-2023-hybrid}. Given that medical tasks are typically ambiguous and not trivial for modern models, HUQ may be a highly suitable approach for uncertainty quantification in such settings.

\paragraph{Our Modification of the Hybrid Method (HUQ-2).}
  We note that the original version of the HUQ method has several shortcomings. In some cases, instances from the ambiguous in-distribution may have greater uncertainty than those from the out-of-distribution area around the decision boundary. Furthermore, it also introduces an issue with ranking across different subsets of the validation set. To address these issues, we propose a modification of HUQ, namely HUQ-2. We employ the squared ranks of instances, weighted with the inverse dependency from another type of uncertainty. This dependency on another type of uncertainty helps to achieve a more precise balance between the types of uncertainty. At the same time, the squared ranks are necessary to prioritize one type of uncertainty for cases in out-of-distribution or ambiguous in-distribution areas. The total uncertainty score according to HUQ-2 is:
  \begin{equation}
    U_{\text{HUQ-2}}(\xv) = (1 - \alpha) R\bigl(U_{\text{E}}(\xv), \DC\bigr)^2 l_1\bigl(U_{\text{A}}(\xv)\bigr) + \alpha R\bigl(U_{\text{A}}(\xv), \DC\bigr)^2 l_2\bigl(U_{\text{E}}(\xv)\bigr),
  \end{equation}
  where $l_i(\uv) = 1 - \frac{R(\uv, \DC)}{c * N}, i\in[1,2]$, $N$ is a number of instances in the validation set, and $c \in [1,2,3]$ is a hyperparameter, which is used to indicate the rate at which the weight will decrease with an increase of the rank. 
  
  Given that we assume that $U_{\text{E}}(\xv)$ is relatively small in the in-distribution area, then we obtain $U_{\text{T}}(\xv) \approx U_{\text{A}}(\xv)$ in this area similarly to the original HUQ approach.

\subsection{Experimental Setup}
  In this section, we overview the experimental design considered in this article. We start by describing the datasets, used models, and evaluation metrics. In the second part, we present and analyze the obtained results.

\subsubsection{Datasets}

\paragraph{MIMIC-III.} 
  In our work, we consider several different medical tasks. The first is a binary classification task for the mortality prediction generated from the MIMIC-III~\cite{Johnson2016-tg} dataset. It consists of hospital admissions collected between the years 2001 and 2012. MIMIC-III includes comprehensive clinical data such as patient demographics, laboratory results, vital signs, caregiver notes, procedures, and discharge summaries and has been widely used for machine learning research in healthcare. For our task, we follow the standard procedure~\cite{wen-etal-2020-medal} of deriving binary labels for mortality prediction.

\paragraph{MIMIC-IV.}
  The second task is multi-label medical code prediction (MCP), generated from the MIMIC-IV~\cite{johnson2023mimic,mimic-4-medical} dataset. MIMIC-IV, an updated version of MIMIC-III, contains hospital admissions from 2008 to 2019, featuring improvements in data accuracy and a shift toward ICD-10 coding. For our specific MCP task, we focus on predicting the top 50 most frequent diagnoses using ICD-10 codes, a standard approach in similar studies, which span a wide range of medical conditions~\cite{Mullenbach2018ExplainablePO, Shi2017TowardsAICD}. MIMIC-IV offers more structured and granular information than MIMIC-III, with enhanced coverage of patient encounters, making it suitable for modern machine learning tasks.

  \begin{table}[]
    \begin{tabular}{@{}lllll@{}}
      \toprule
                                        & \textbf{OV Dataset} & \textbf{MIMIC-IV Top-50} & \textbf{MIMIC-III} &  \\ \midrule
      Number of patients                & 1,133,811           & 61,265           & 34,692     &  \\
      Number of records                 & 9,000,988           & 115,250          & 260,195    &  \\
      Number of unique classes          & 363                 & 50               & 2          &  \\
      Avg. number of codes / record     & 1                   & 5                & 1          &  \\
      Avg. number of words / record     & 475                 & 1,587            & 246        &  \\
      Avg. number of records / patient  & 8                   & 2                & 7          &  \\ \bottomrule
    \end{tabular}
    \caption{Mortality and ICD codes prediction datasets descriptive statistics.}
    \label{tab:datastats}
  \end{table}

  \begin{table}[]
\centering
\begin{tabular}{@{}lllllll@{}}
\toprule
                                  & \textbf{DE} & \textbf{DSM} & \textbf{AL} & \textbf{AD} & \textbf{AC} & \\ \midrule
Number of patients (records)               & 557           & 224     & 202 & 190 & 413            &  \\
Number of unique classes          & 2              & 2         & 2 & 2 & 2             &  \\
Avg. number of words / record & 325            & 289       & 153 & 91 & 14            &  \\
 \bottomrule
\end{tabular}
\caption{Descriptive statistics for datasets for mental disorders detection. Each patient has only one record.}
\label{tab:psy_datastats}
\end{table}

\paragraph{Outpatient Visits (OV).}
  The last is the multi-class medical code prediction task, derived from the private healthcare data source named \textit{Outpatient Visits (OV) dataset}, which comprises anonymized health records collected from the Central Medical Information and Analysis System of a large European city between 2017 and 2021. Each record includes patient demographics, outpatient doctor visits (complaints, medical history, diagnoses with ICD codes), laboratory results (test names, results, and reference values), and instrumental examination details (protocols and conclusions). We use the same preprocessing pipeline for all three described datasets to ensure consistency and fair comparison across tasks~\cite{wen-etal-2020-medal} and provide their detailed statistics in Table~\ref{tab:datastats}.

  % Throughout the dataset correction, the target ICD diagnoses subset underwent three stages, starting with 256 diagnoses, then expanding to 571, and finally narrowing to 363 relevant diagnoses based on specific exclusion criteria.
  % The ICD diagnosis subset was refined in three stages: first, 256 common diagnoses were selected; then, 571 were expanded to cover 95\% of cases; and finally, 363 diagnoses were narrowed to 363 by excluding those irrelevant to general therapy or clinical recommendations. This process ensured that the final subset was clinically focused.
  
\paragraph{Mental Disorder Detection.}
  In addition to the aforementioned tasks, we consider five datasets for detecting mental disorders. Two datasets are targeted at depression detection, while three others are aimed at anxiety detection. The first dataset, Depression-Essays (DE; \cite{stankevich2019predicting}), contains essays, part of which are written by subjects with clinically diagnosed depression. The second one, Depression-Social Media (DSM; \cite{ignatiev2022predicting}), consists of text messages from a social network with targets based on the Beck Depression Inventory~\cite{beck1996beck} questionnaire. The preprocessing pipeline for this dataset and all mental disorder detection datasets employs the approach from~\cite{kuzmin2024mentaldisordersdetectionera}. The Anxiety-Letter (AL) and Anxiety-Description (AD) datasets are based on the RusNeuroPsych corpus~\cite{litvinova2018rusneuropsych} and employ Hospital Anxiety and Depression Scale~\cite{bjelland2002validity} to obtain target variables. In the AL dataset, subjects were asked to write an informal letter to a friend, while in AD, the same subjects described a picture. Finally, the Anxiety-COVID dataset (AC; \cite{medvedeva2021lexical}) uses SCL-90-R~\cite{Derogatis1983SCL90RAS} questionnaire to determine the anxiety level of subjects, while subjects were asked to write a commentary about the pandemic and self-isolation. Labels for AL, AD, and AC datasets were built according to the same procedure as in~\cite{kuzmin2024mentaldisordersdetectionera}. Table~\ref{tab:psy_datastats} shows the statistics for all these datasets.

\paragraph{Dataset Selection Rationale.}
  The selection of datasets is based on a strategy that covers both standard medical tasks and specific scenarios, contributing to a more comprehensive investigation. The MIMIC-III and MIMIC-IV datasets are well-established benchmarks for research in healthcare machine learning, providing tasks related to mortality prediction and multi-class classification based on widely used and validated clinical data~\cite{Johnson2016-tg,johnson2023mimic}. The OV dataset also considers a multi-class diagnosis classification. It complements the MIMIC data with a more narrowly focused example, providing real-world collected data reflecting clinical practice's complexity and diversity. The inclusion of additional datasets related to depression and anxiety disorders (DE, DSM, AL, AD, AC) broadens the range of tasks to include multi-label classification in the field of mental health, adding diversity to the study and deepening the analysis. This approach allows us to generalize the findings to various medical contexts and thoroughly investigate uncertainty quantification methods in complex, real-world scenarios.

\subsubsection{Models}
  We conduct experiments with state-of-the-art Transformer-based models~\cite{vaswani2023attentionneed} in the medical domain. Specifically, we utilize GatorTron-base~\cite{yang2022large} for mortality detection in the MIMIC dataset, Clinical-Longformer~\cite{clinical-longformer} for medical code prediction in the MIMIC dataset, and Longformer~\cite{longformer} for medical code prediction in the OV dataset. For the mental disorders detection tasks, we used the RuBioRoBERTa model~\cite{yalunin2022rubioroberta}. Each model was fine-tuned for the specified task. The training hyperparameters are presented in Table~\ref{tab:hp} of Appendix~\ref{appendix:hp}.

\subsubsection{Metrics}
  We utilize the standard metrics to evaluate uncertainty quantification methods for the selective prediction task~\cite{selective-classification}. In this task, the model abstains from predictions with the highest uncertainty scores, and rejected instances are removed from the dataset to other alternate procedures. Therefore, to evaluate the performance of the methods in this task in mortality prediction and multi-class medical diagnosis, we use the standard metric: area under the risk coverage curve (RC-AUC; \cite{rc-auc}). 

  The predictions in a dataset are sorted in ascending order by uncertainty to abstain from some percentage of the most uncertain predictions. The percentage of predictions that are not removed is called a coverage rate, and the total loss of the remaining predictions is called the selective risk. The curve obtained when varying the rejection threshold is called the risk coverage (RC) curve and demonstrates a dependence of the selective risk from the coverage rate. Finally, the RC-AUC represents a cumulative sum of the selective losses for each coverage rate. Lower absolute values of RC-AUC indicate better performance.

  \begin{equation}
  \label{eq:rc_auc}
    \begin{cases}
      \gamma(\theta) = \frac{1}{|\DC|}\sum_{(\xv_i, y_i) \in \DC} {q_i}, \\
      r(\theta) = \frac{\sum_{(\xv_i, y_i) \in \DC} {q_i} l_i}{\sum_{(\xv_i, y_i) \in \DC} {q_i}},
    \end{cases}
  \end{equation}
  where $q_i = \mathbf{1}[U(\xv_i) < \theta]$, binary loss $l_i = L(h(\xv_i), y_i)$,    $R(U, \DC) = \sum_{\xv_i \in \DC} \mathbf{1}[U(\xv_i) < U]$. 
  
  However, this metric does not apply to the multi-label classification task since it uses a binary loss function in practice. Therefore, we measure the area under the multi-label task's F1-micro rejection curve (FR-AUC). This metric is calculated similarly but computes the F1-micro of the remaining predictions for the specified coverage rate. Higher values of FR-AUC indicate better performance. 

  The absolute values of RC-AUC and FR-AUC are unnormalized, making analyzing these metrics challenging. Moreover, RC-AUC is meaningful only when comparing methods, while its absolute value for a single method is not informative. Following previous work~\cite{malinin2020uncertainty}, to enhance the interpretability, we normalize both metrics using the AUC for the random uncertainty scores and the oracle (the best possible) scores: 
  \begin{equation}
    \text{AUC}_{norm} = \frac{\text{AUC} - \text{AUC}_{rand}}{\text{AUC}_{oracle} - \text{AUC}_{rand}}.
  \end{equation}
  A rejection curve for the oracle scores goes linearly up to a quality of 1.0, representing the optimal increase rate. A rejection curve for the random uncertainty scores for all points is close to the base model quality, representing the random rejection order. A $\text{AUC}_{norm}$ of 1.0 indicates optimal rejection, while a value of 0.0 indicates random rejection.

  Finally, for both metrics, we calculate the area under the entire curve and the first 50\% of the curve. In most applications, we are interested in rejecting only the predictions with the highest uncertainty values, and rejecting more than 50\% incoming objects is rarely an option. That is why the part of the rejection curve corresponding to higher rejection rates is less informative for practitioners, and we focus on the area under the first 50\% of the curve. 

  The case of multi-label classification should be additionally discussed due to the peculiarities of this problem. It is natural to expect the model to be confident in some diagnoses while uncertain in others. Thus, one may consider a task of selective prediction where one rejects to predict not for the instances but only the particular labels of these instances. The FR-AUC can still be computed based on the F1-micro metric for the remaining label-instance pairs. We call such a selective prediction approach \textit{label-wise selective prediction} instead of the standard \textit{instance-wise selective prediction}.

\subsubsection{Implementation Details}
  In our experiments with density-based and hybrid methods, we use the following pipeline:
  \begin{enumerate}
    \item Extract embeddings from the penultimate layer of the trained model on the training and validation datasets. We use the entire training dataset for the MIMIC and Mental Disorders datasets, while for the OV dataset, we use a subsample of 300 thousand instances.

    \item Fit the parameters of the density-based methods using the extracted embeddings from the training dataset.

    \item On the validation dataset, compute SR and density-based uncertainty scores. Using these scores, we fit the hyperparameters of the HUQ and HUQ-2 methods, which optimize RC-AUC or FR-AUC depending on the task.

    \item During the inference for the test instance, we use the fitted parameters and hyperparameters for density-based and HUQ methods for calculating the final uncertainty scores.  
  \end{enumerate}
  To ensure the results' robustness and provide standard deviation, we train five models with different random seeds for all datasets except the OV dataset. Due to its large size, we trained only a single model for the OV dataset. We use 20 stochastic forward passes for each test instance for the Monte Carlo dropout methods.

\section{Data Availability}
  For the tasks of mortality prediction and ICD-code assignment, we utilized two publicly available datasets: MIMIC-III\footnote{\url{https://physionet.org/content/mimiciii/1.4/}}, and MIMIC-IV\footnote{\url{https://physionet.org/content/mimiciv/3.1/}}. Both datasets require a formal application process, including a data use agreement and CITI training on human research protection. Additionally, we used a private Outpatient Visits (OV) dataset that is not publicly available due to institutional restrictions on data privacy, with access limited to authorized researchers under strict ethical and confidentiality agreements. The same limitations apply to datasets for the detection of mental disorders.

\section{Code Availability}
  The source code and scripts necessary to reproduce the experiments presented in this paper are available online\footnote{\url{https://anonymous.4open.science/r/medical_uncertainty_quantification-64E7/}}.

% bibliography
  \bibliographystyle{naturemag}
  \bibliography{references}

\section*{Author contributions}
  A.V., A.P., A.N., A.S., and M.P. were involved in conceptualization, study design and methodology. A.V., I.S., G.K., and A.N. were responsible for data extraction and preparation. A.V., A.B. and G.K. executed machine learning experiments and performed data analysis. All authors contributed to writing the original draft of the manuscript. A.N., A.S. and M.P. provided supervision for the project. All authors have read and approved the manuscript.

\section*{Competing interests}
  The authors declare no competing interests.

\appendix
\newpage

\section{Additional Experimental Results}
\label{appendix:results}
  In this section, we present some additional experimental results that complement those in the main part of the paper. 

\subsection{Results for Full Rejection Curves}
\begin{figure*}[ht]
    \footnotesize
    \centering
    \begin{minipage}[h]{0.4\linewidth}
    \center{\includegraphics[width=4.4cm]{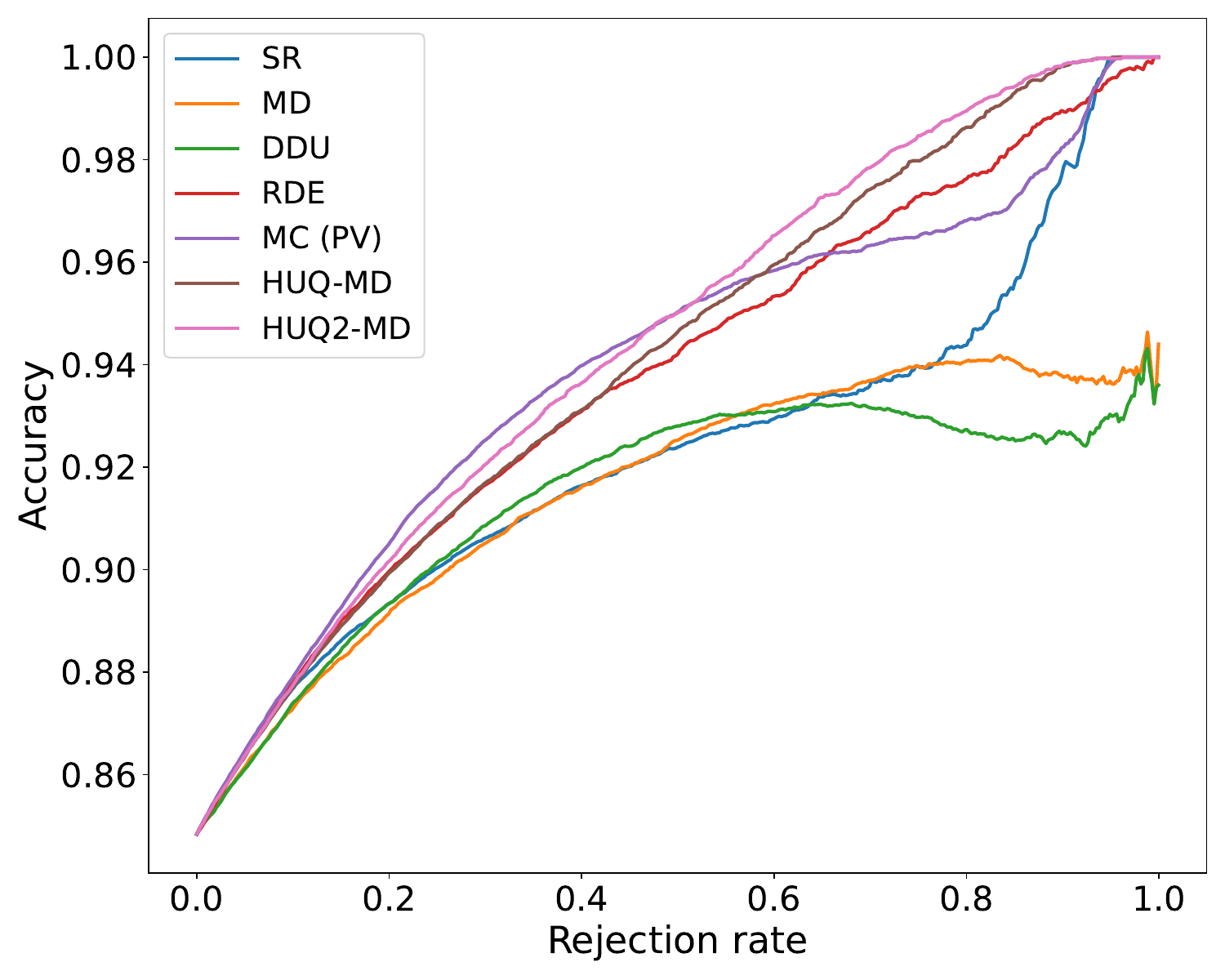} \\
    a) MIMIC-III mortality prediction 
    }
    \end{minipage}
    \begin{minipage}[h]{0.4\linewidth}
    \center{\includegraphics[width=4.4cm]{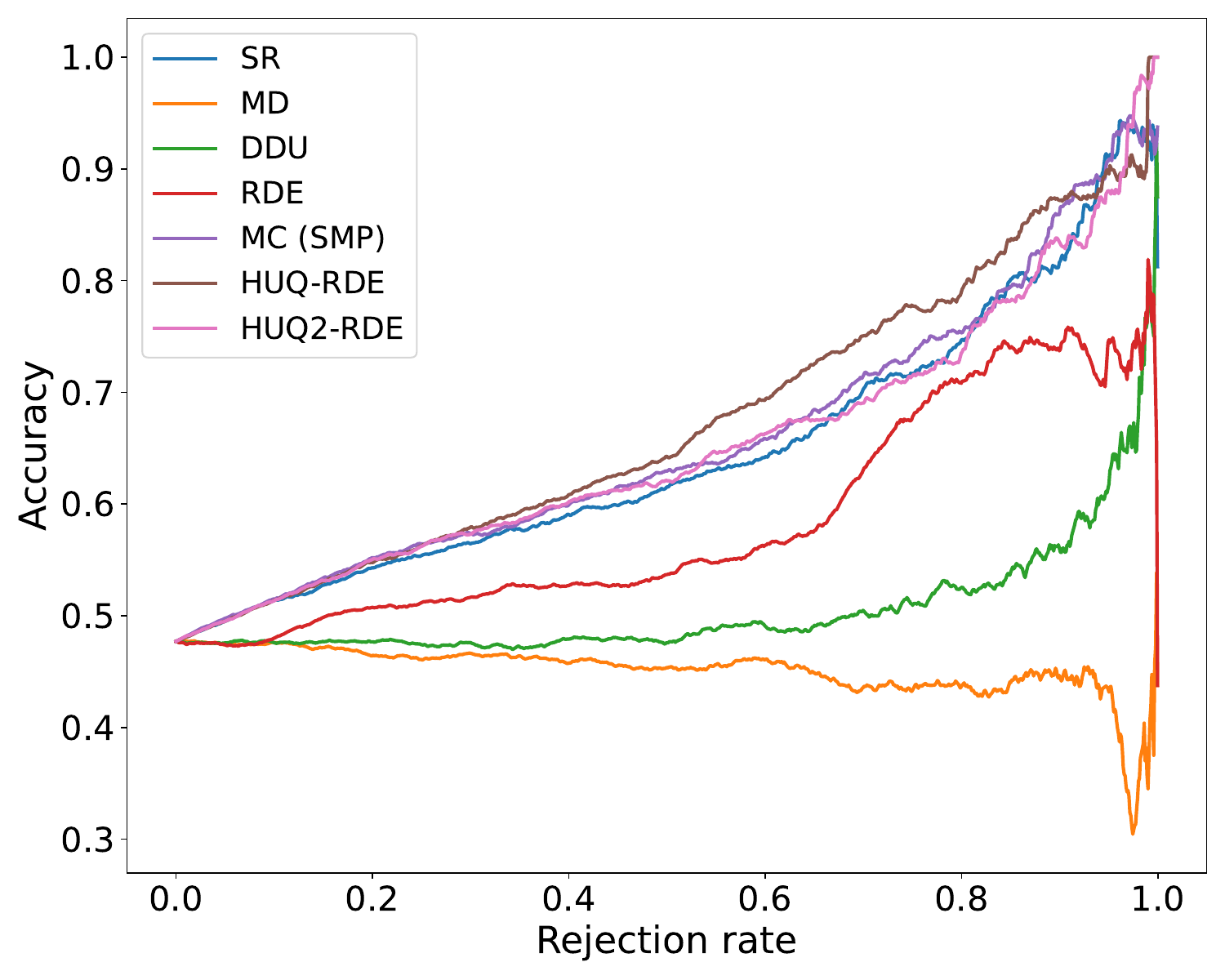} \\
    b) OV medical code prediction
    }
    \end{minipage}
    \begin{minipage}[h]{0.4\linewidth}
    \center{\includegraphics[width=4.4cm]{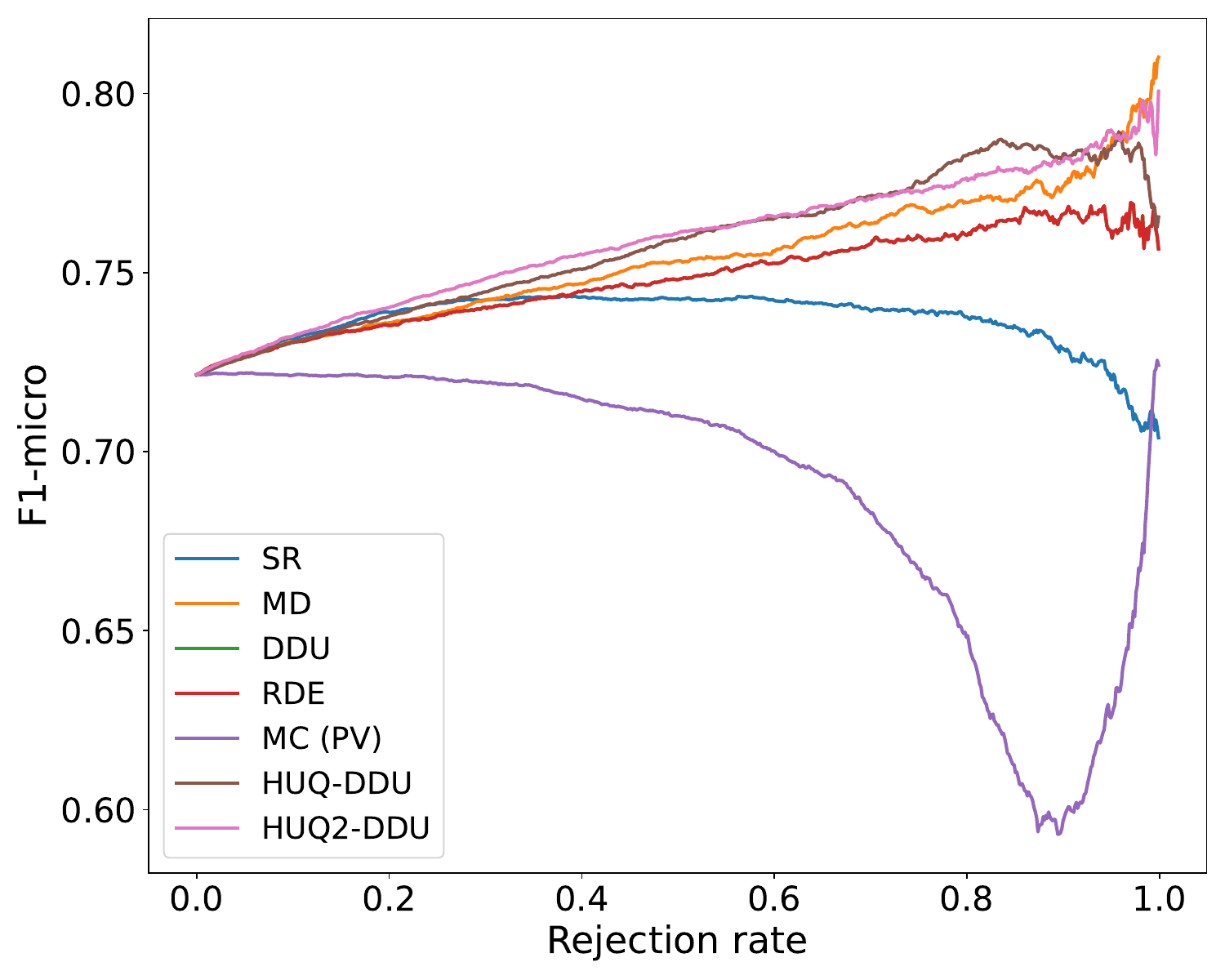} \\
    c) MIMIC-IV medical code prediction
    }
    \end{minipage}
    \caption{Full rejection curves for the selected methods for the considered tasks.}
    \label{fig:full_rejection_curves}
  \end{figure*}

\begin{figure*}[t]
    \footnotesize
    \centering
    \begin{minipage}[h]{0.45\linewidth}
    \center{\includegraphics[width=5.5cm]{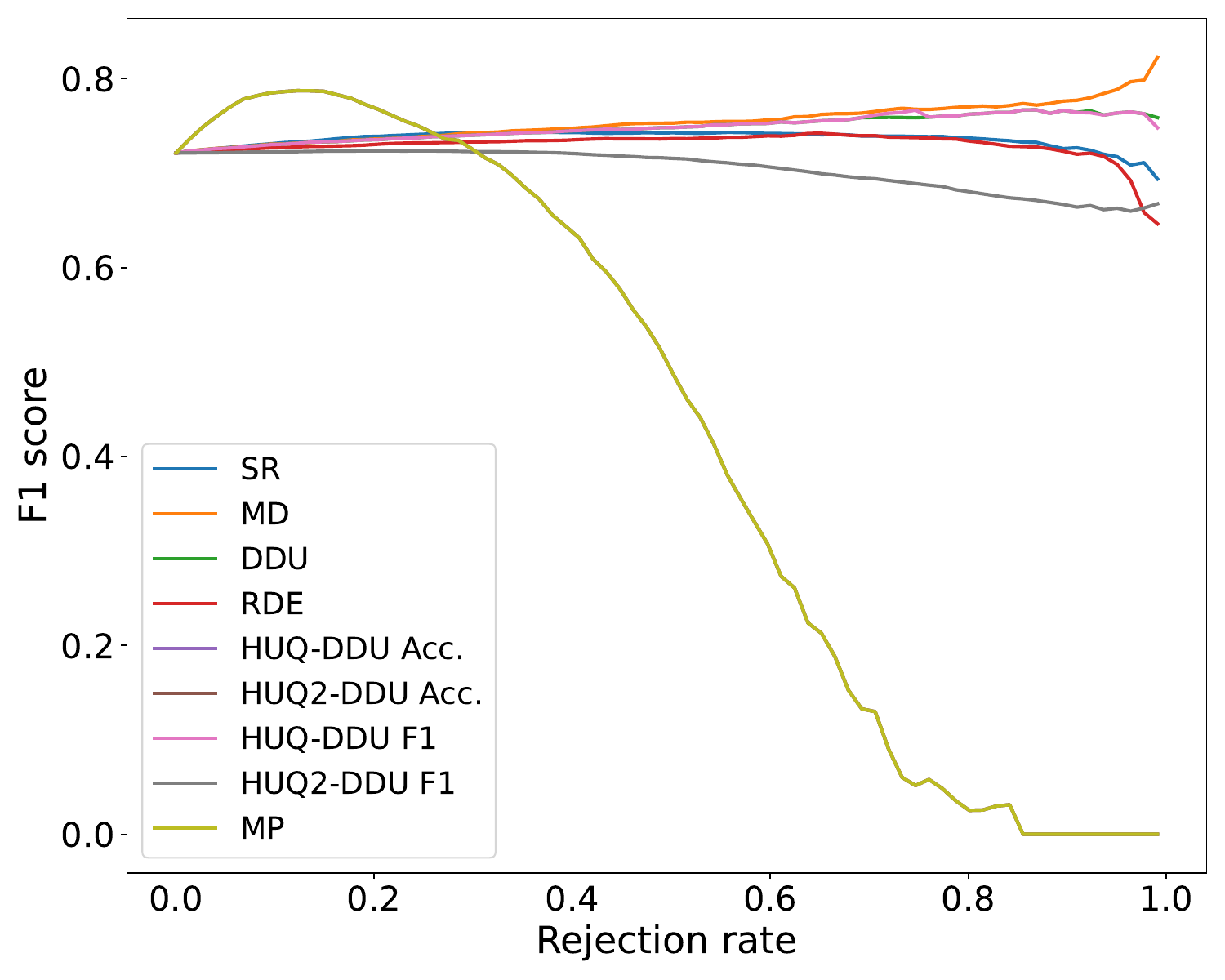} \\%{images/huq_label_wise/mimic_mcp_acc_flatten.pdf} \\
    % b) Accuracy
    }
    \end{minipage}
    \begin{minipage}[h]{0.45\linewidth}
    \center{\includegraphics[width=5.5cm]{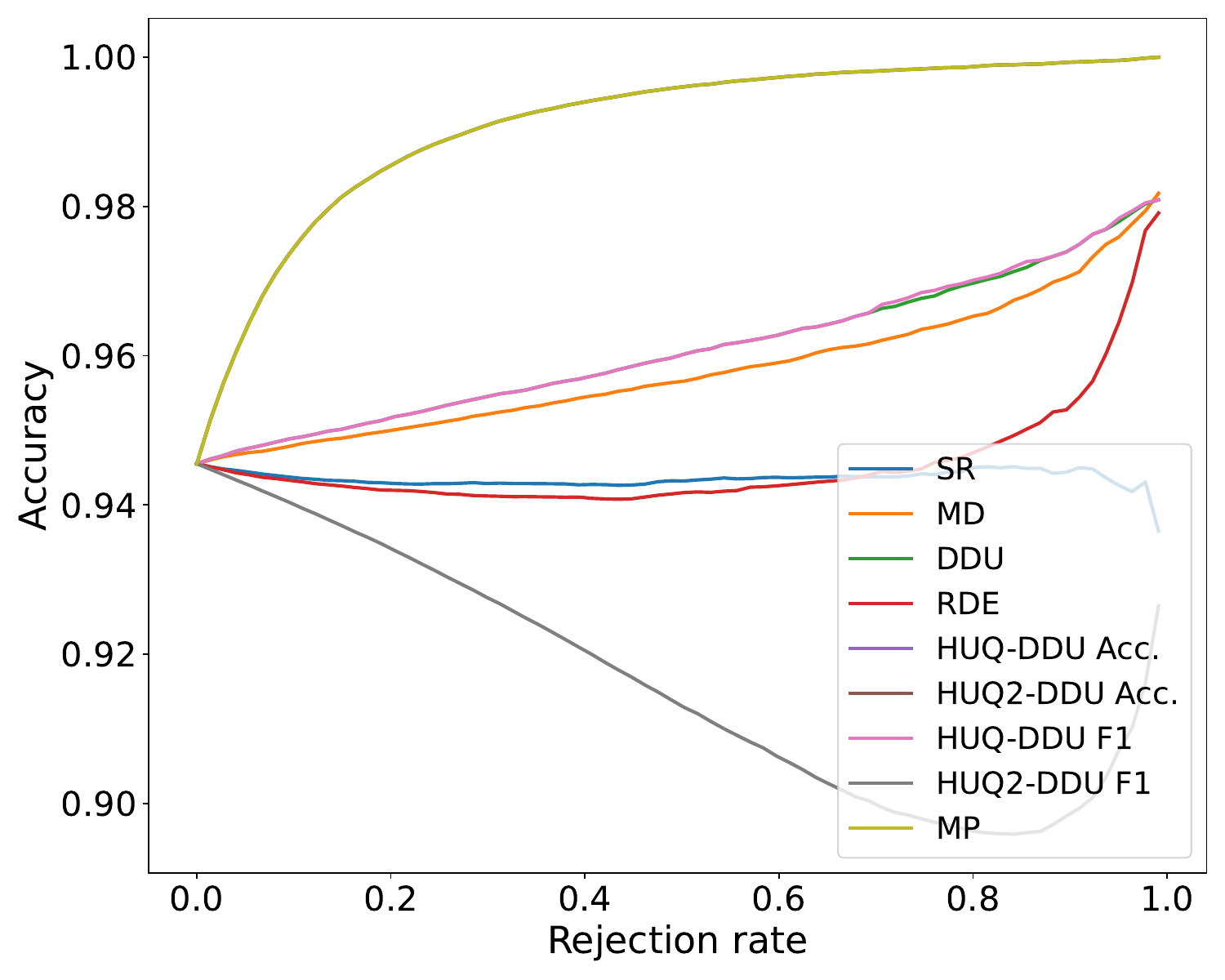} \\%{images/huq_label_wise/mimic_mcp_flatten_f1_huq_on_acc.pdf} \\
    % a) F1 score 
    }
    \end{minipage}
    \caption{\label{fig:full_label_rejection_curves} Full rejection curves for the selected methods for the MIMIC medical code prediction task for general rejection vs label-wise approach. Due to the selected hyperparameters on the validation set, the HUQ and HUQ-2 methods overlap with the MP, MD, and DDU methods.% HUQ hyperparameters are fitted using the accuracy rejection metric.
    }
  \end{figure*}

  In the main part of the paper, we present the results only for rejection of up to 50\% of test points. We proceed in this way as, in practical scenarios, it is unlikely that one wants to reject large fractions of points. Additionally, the results for high rejection rates are less robust as they are based on a relatively small number of points. In Figure~\ref{fig:full_rejection_curves}, we present full rejection curves for MIMIC-III mortality prediction, OV medical code prediction, and MIMIC-IV medical code prediction tasks. The corresponding AUC values are shown in Table~\ref{tab:mimic_100}. Additionally, we plot full rejection curves for the experiments with label-wise approaches in Figure~\ref{fig:full_label_rejection_curves}. For the HUQ and HUQ-2 methods, we consider calibration to be either the F1 score or accuracy and present results for both variants.

  \begin{table*}[!ht] \centering%\resizebox{0.9\textwidth}{!}{
\begin{tabular}{l|c|c|c}
\toprule
\textbf{UQ Method} & \textbf{MIMIC Mortality} & \textbf{OV Dataset MCP} & \textbf{MIMIC MCP} \\
\midrule
SR & 0.55$\pm$0.01 & 0.48 & 0.10$\pm$0.01 \\
Entropy & -0.75$\pm$0.02 & 0.47 & 0.12$\pm$0.01 \\
Delta & -0.75$\pm$0.02 & 0.47 & 0.12$\pm$0.00 \\
Beta & 0.53$\pm$0.03 & 0.48 & -0.30$\pm$0.01 \\\midrule
MC (PV) & 0.66$\pm$0.01 & 0.46 & -0.22$\pm$0.01 \\
MC (SMP) & 0.64$\pm$0.01 & \underline{0.51} & -0.26$\pm$0.01 \\
MC (BALD) & 0.67$\pm$0.01 & 0.47 & -0.21$\pm$0.01 \\\midrule
NUQ ep. & 0.65$\pm$0.03 & 0.18 & 0.11$\pm$0.01 \\
DDU & 0.47$\pm$0.03 & 0.08 & 0.20$\pm$0.03 \\
RDE & 0.65$\pm$0.02 & 0.3 & 0.06$\pm$0.01 \\
MD & 0.49$\pm$0.03 & -0.07 & 0.24$\pm$0.02 \\\midrule
HUQ-DDU & \underline{0.67$\pm$0.01} & \underline{0.48} & \underline{0.27$\pm$0.01} \\
HUQ2-DDU & \underline{0.69$\pm$0.01} & 0.48 & \textbf{0.28$\pm$0.01} \\
HUQ-RDE & 0.65$\pm$0.02 & \textbf{0.56} & 0.12$\pm$0.01 \\
HUQ2-RDE & 0.66$\pm$0.02 & \underline{0.5} & 0.11$\pm$0.03 \\
HUQ-MD & \underline{0.67$\pm$0.01} & 0.48 & \underline{0.26$\pm$0.02} \\
HUQ2-MD & \textbf{0.69$\pm$0.01} & 0.48 & \underline{0.27$\pm$0.01} \\
\bottomrule
\end{tabular}
%}
\caption{\label{tab:mimic_100} Results for the selective classification task for MIMIC mortality, OV, and MIMIC MCP  datasets. For the MIMIC mortality detection and OV datasets, we use normalized RC-AUC$\uparrow$ on the full curve. For the MIMIC medical code prediction task, we use the area under the full F1-micro rejection curve (FR-AUC$\uparrow$). The best results for each dataset are shown in bold. We underline top-3 methods after the best.}
\end{table*}

  \newpage
  \clearpage

\subsection{Rejection Curves for Mental Disorder Datasets}
\begin{figure*}[ht]
    \footnotesize
    \centering
    \begin{minipage}[h]{0.4\linewidth}
    \center{\includegraphics[width=4.4cm]{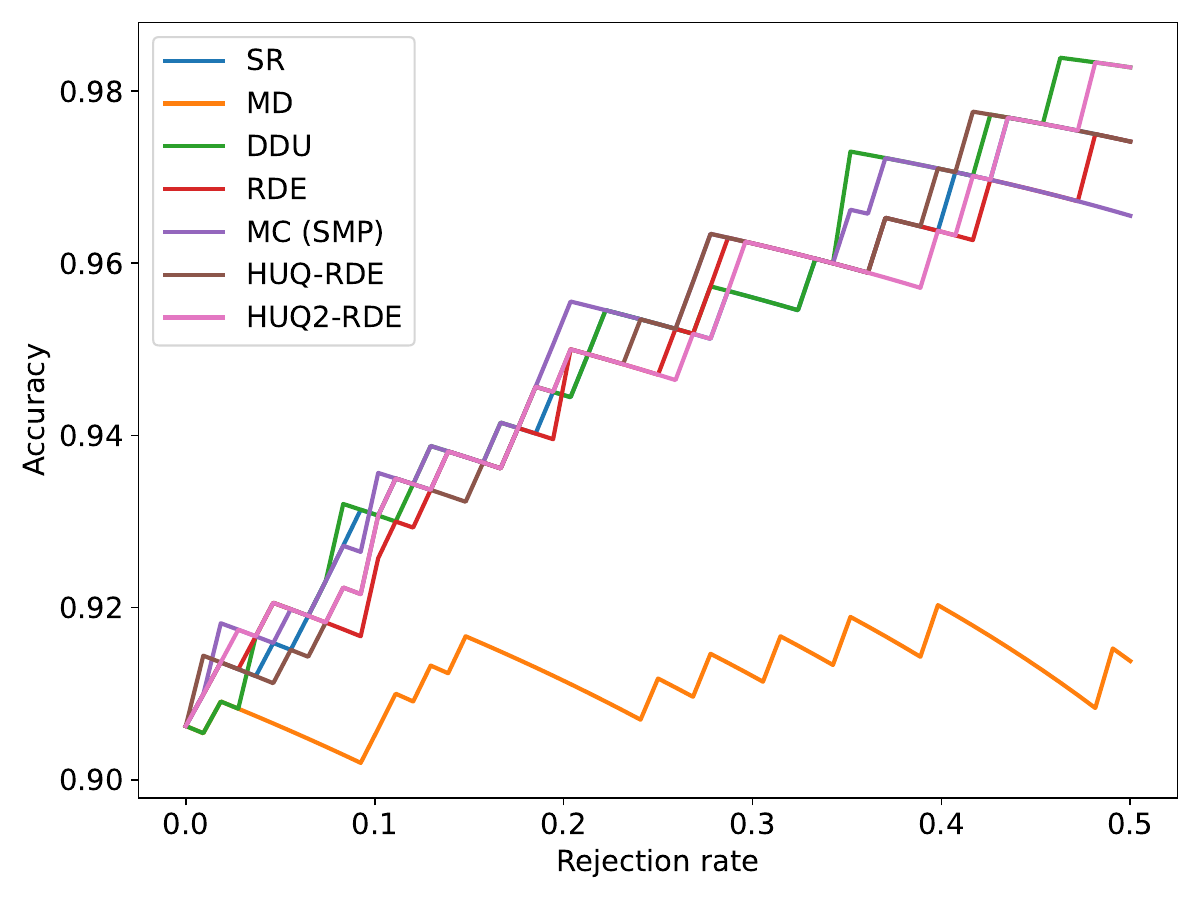} \\
    a) DE 
    }
    \end{minipage}
    \begin{minipage}[h]{0.4\linewidth}
    \center{\includegraphics[width=4.4cm]{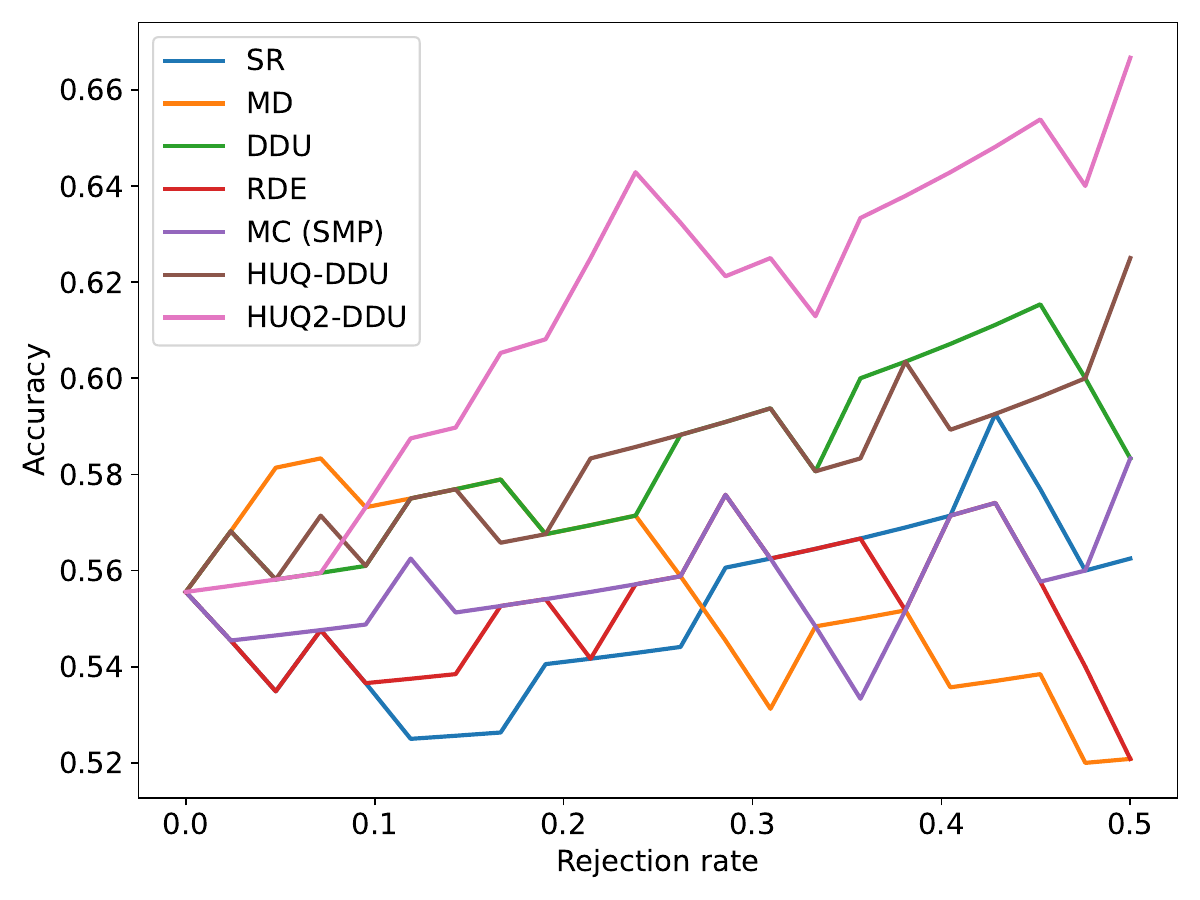} \\
    b) DSM 
    }
    \end{minipage}
    \begin{minipage}[h]{0.4\linewidth}
    \center{\includegraphics[width=4.4cm]{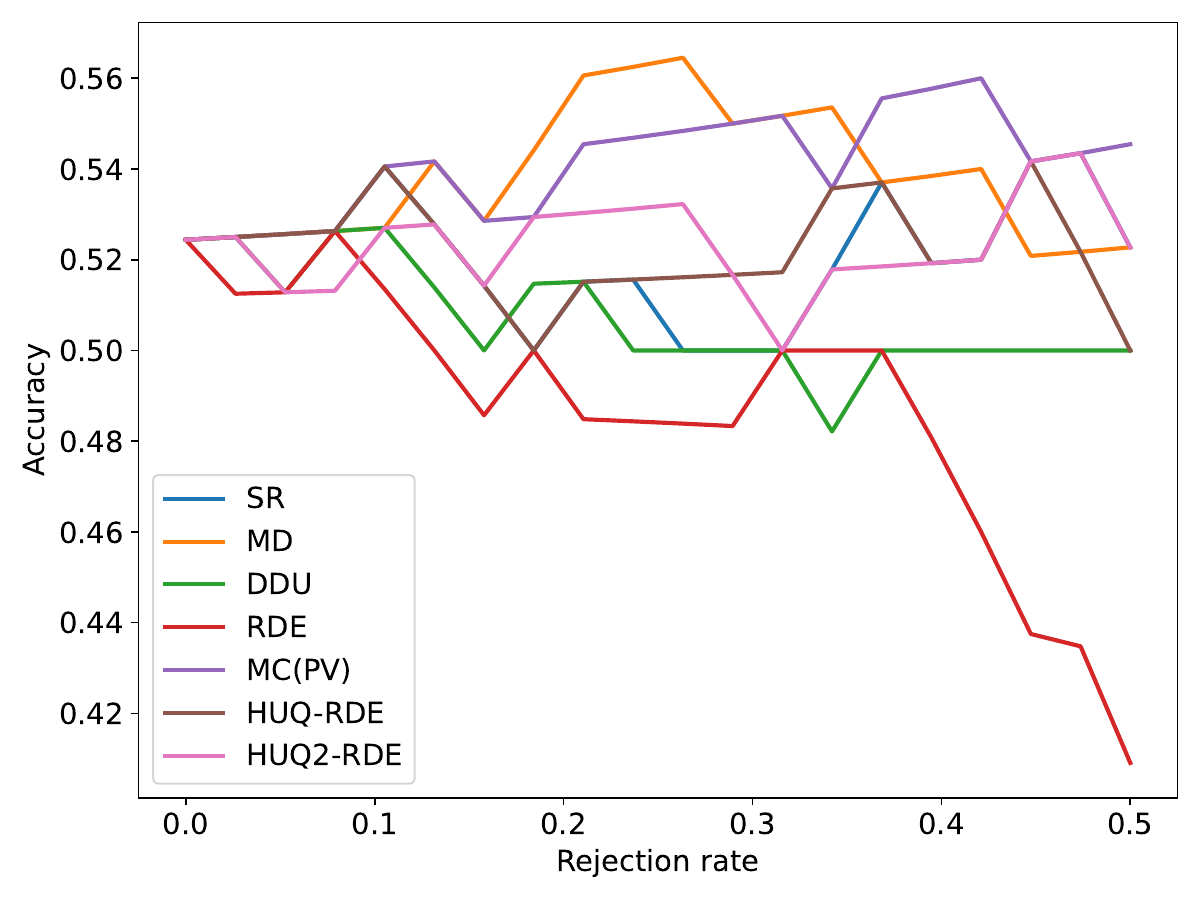} \\
    a) AL 
    }
    \end{minipage}
    \begin{minipage}[h]{0.4\linewidth}
    \center{\includegraphics[width=4.4cm]{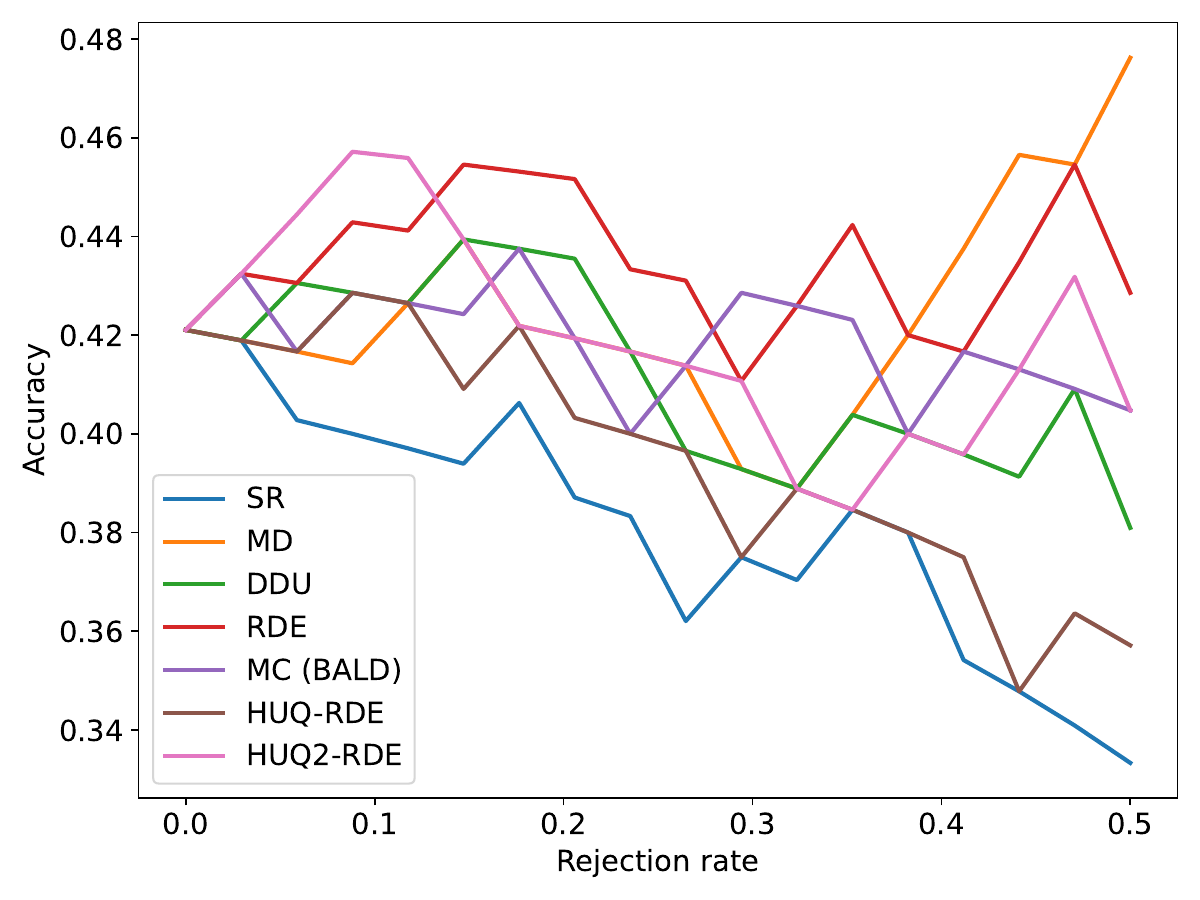} \\
    b) AD 
    }
    \end{minipage}
    \begin{minipage}[h]{0.4\linewidth}
    \center{\includegraphics[width=4.4cm]{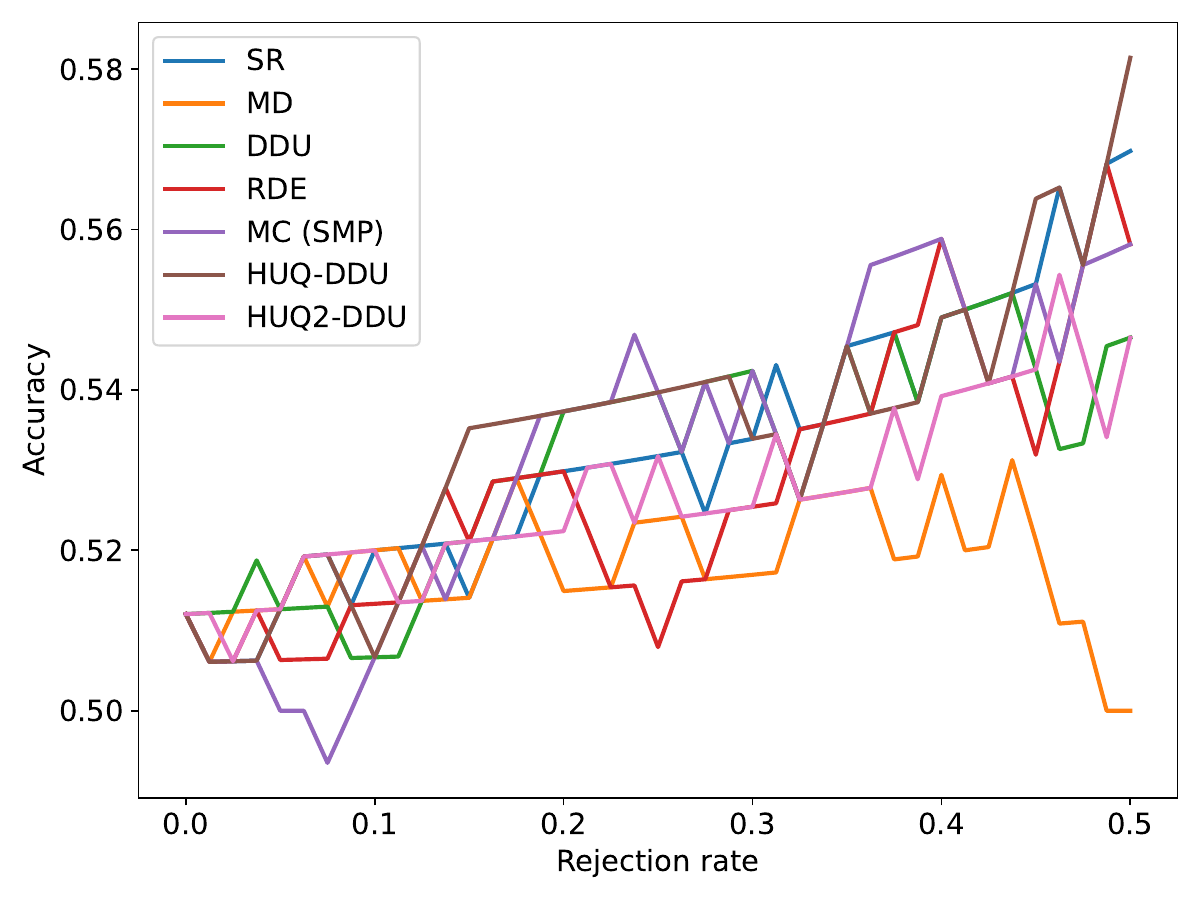} \\
    c) AC
    }
    \end{minipage}
    \caption{Rejection curves for the selected methods for the mental disorders detection tasks. The curves are noisy due to the relatively small size of the test set.}
    \label{fig:rejection_curves_psy}
  \end{figure*}

\begin{figure*}[ht]
    \footnotesize
    \centering
    \begin{minipage}[h]{0.4\linewidth}
    \center{\includegraphics[width=4.4cm]{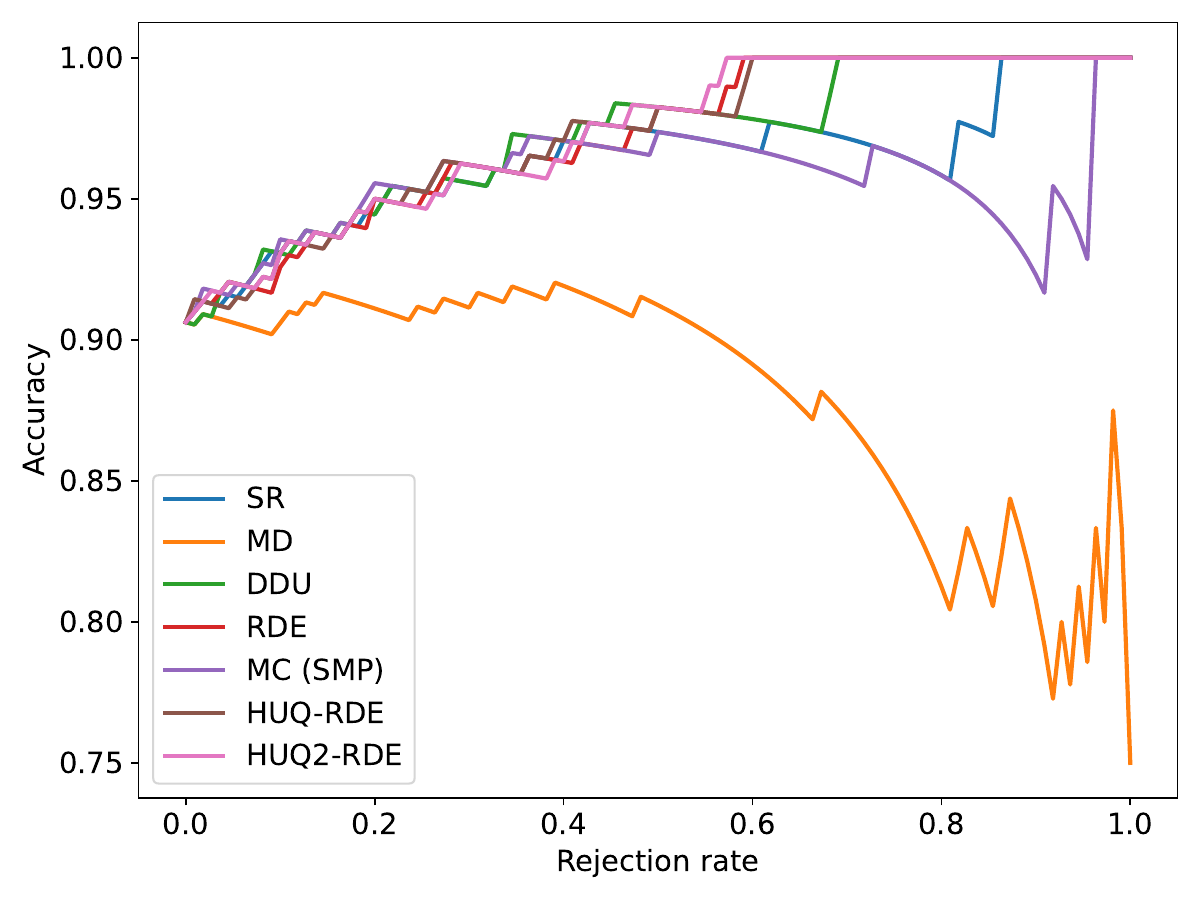} \\
    a) DE 
    }
    \end{minipage}
    \begin{minipage}[h]{0.4\linewidth}
    \center{\includegraphics[width=4.4cm]{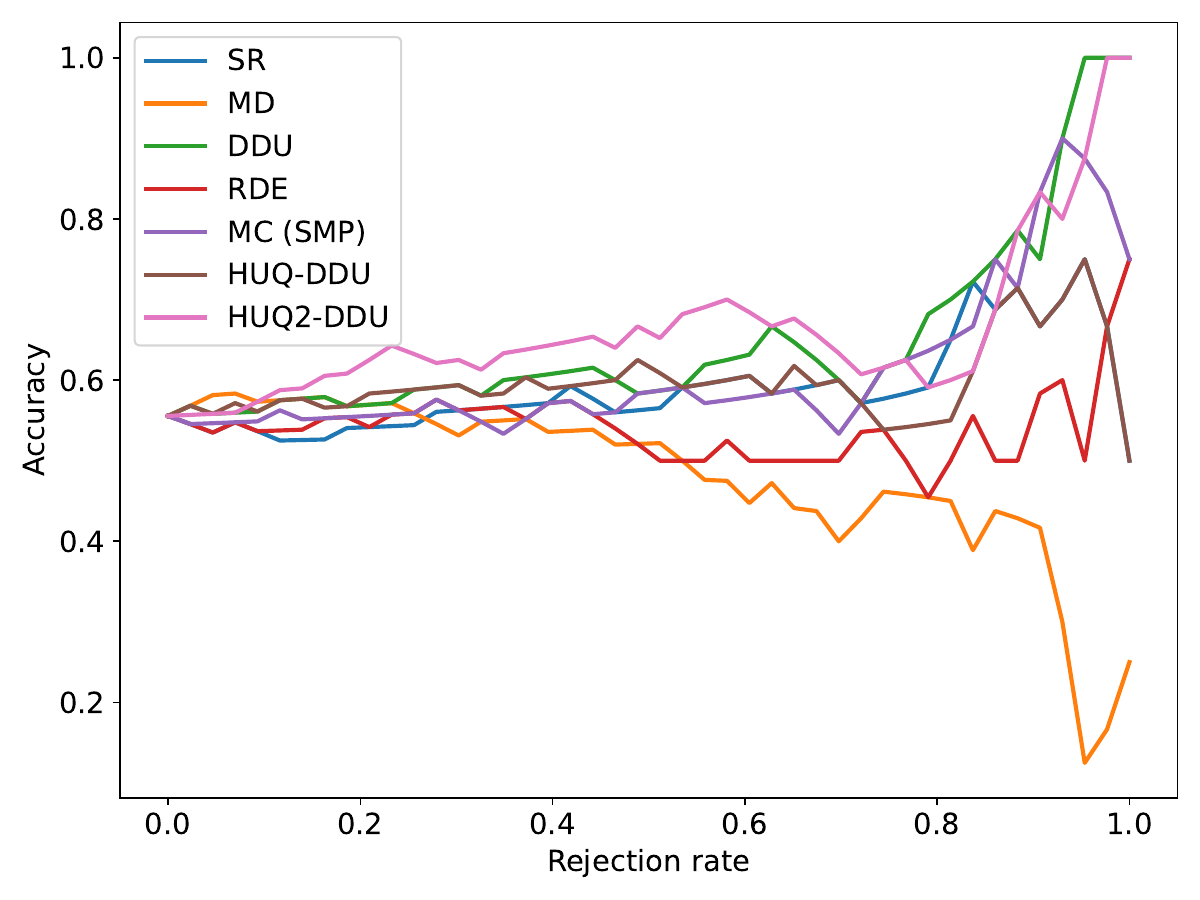} \\
    b) DSM 
    }
    \end{minipage}
    \begin{minipage}[h]{0.4\linewidth}
    \center{\includegraphics[width=4.4cm]{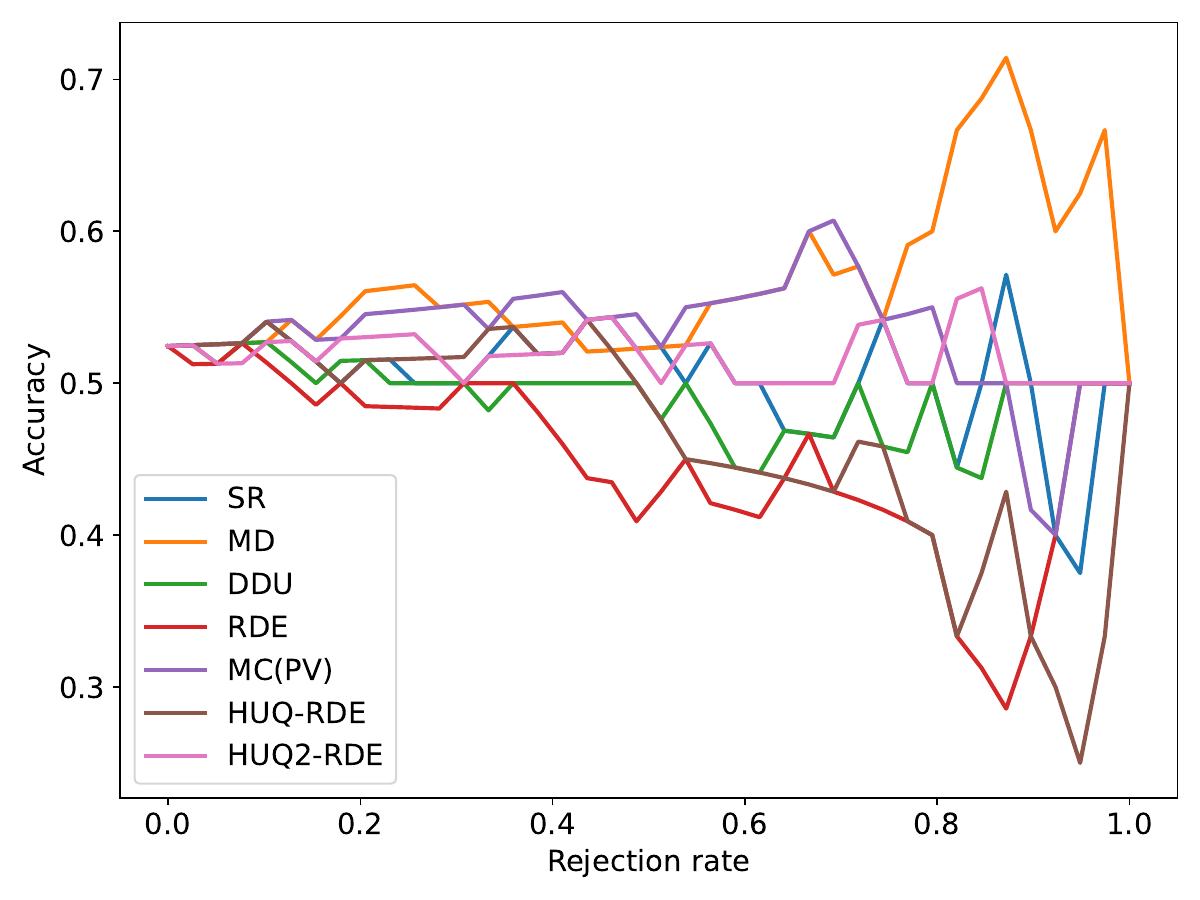} \\
    a) AL 
    }
    \end{minipage}
    \begin{minipage}[h]{0.4\linewidth}
    \center{\includegraphics[width=4.4cm]{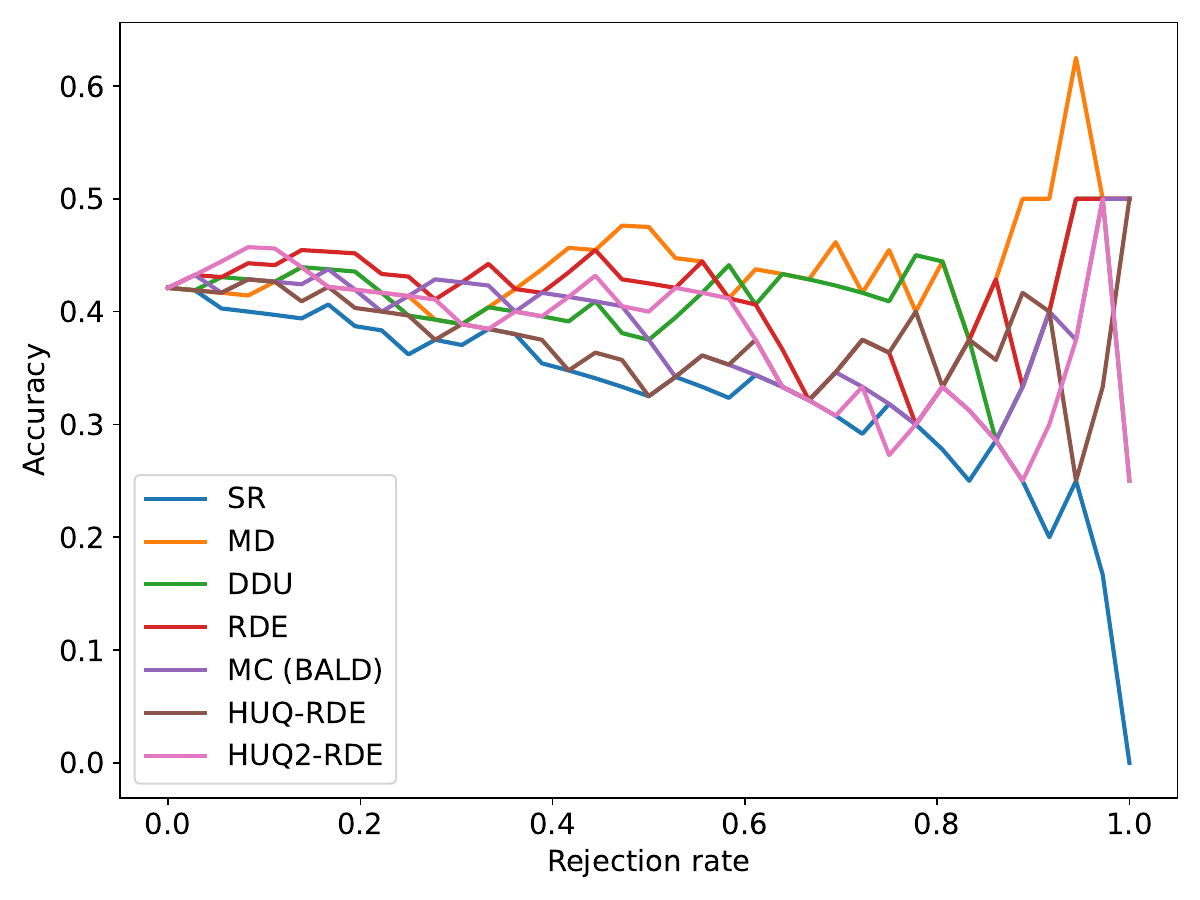} \\
    b) AD 
    }
    \end{minipage}
    \begin{minipage}[h]{0.4\linewidth}
    \center{\includegraphics[width=4.4cm]{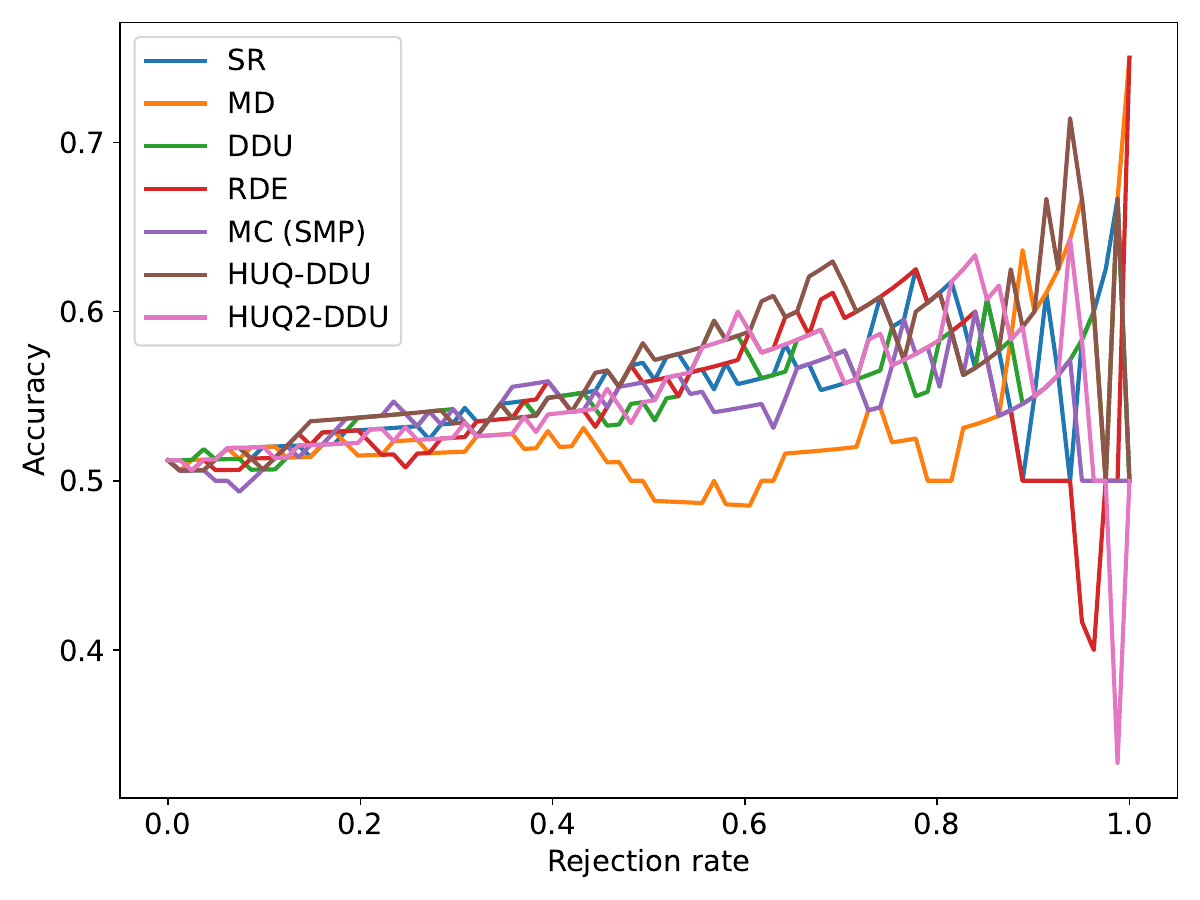} \\
    c) AC
    }
    \end{minipage}
    \caption{Rejection curves for the selected methods for the mental disorders detection tasks. The curves are noisy due to the relatively small size of the test set.}
    \label{fig:full_rejection_curves_psy}
  \end{figure*}

  In this section, we present the rejection curves for mental disorder datasets, see Figure~\ref{fig:rejection_curves_psy} for rejection rates up to 50\%, while full rejection curves are shown in Figure~\ref{fig:full_rejection_curves_psy}. The curves are noisy due to the relatively small size of the test set. The AUC values corresponding to full rejection curves are shown in Table~\ref{tab:psy_ue_rubioroberta_100}.

  \begin{table*}[!ht]
\centering%\resizebox{0.9\textwidth}{!}{
\begin{tabular}{l|c|c|c|c|c}
\toprule
\textbf{UQ Method} & \textbf{DE} & \textbf{DSM} & \textbf{AL} & \textbf{AD} & \textbf{AC} \\
\midrule

SR & 0.62$\pm$0.26 & 0.16$\pm$0.28 & 0.06$\pm$0.29 & -0.16$\pm$0.28 & 0.17$\pm$0.15 \\
Entropy & 0.62$\pm$0.26 & 0.16$\pm$0.28 & 0.06$\pm$0.29 & -0.16$\pm$0.28 & 0.17$\pm$0.15 \\
Delta & 0.62$\pm$0.26 & 0.16$\pm$0.28 & 0.06$\pm$0.29 & -0.16$\pm$0.28 & 0.17$\pm$0.15 \\
Beta & -0.09$\pm$0.64 & 0.07$\pm$0.21 & 0.10$\pm$0.17 & -0.11$\pm$0.28 & 0.03$\pm$0.26 \\ \midrule
MC (PV) & 0.39$\pm$0.48 & 0.17$\pm$0.29 & 0.05$\pm$0.31 & -0.04$\pm$0.20 & 0.05$\pm$0.14 \\
MC (SMP) & 0.42$\pm$0.46 & \underline{0.19$\pm$0.26} & -0.01$\pm$0.26 & -0.04$\pm$0.24 & 0.15$\pm$0.17 \\
MC (BALD) & 0.31$\pm$0.54 & 0.19$\pm$0.27 & 0.03$\pm$0.32 & -0.03$\pm$0.21 & 0.03$\pm$0.13 \\ \midrule
NUQ ep. & 0.49$\pm$0.24 & 0.15$\pm$0.18 & \underline{0.11$\pm$0.24} & -0.02$\pm$0.12 & 0.17$\pm$0.16 \\
DDU & 0.67$\pm$0.18 & \underline{0.29$\pm$0.24} & 0.03$\pm$0.19 & \underline{0.09$\pm$0.24} & 0.15$\pm$0.14 \\
RDE & \underline{0.68$\pm$0.20} & -0.01$\pm$0.20 & -0.08$\pm$0.15 & \underline{0.05$\pm$0.15} & \underline{0.18$\pm$0.22} \\
MD & -0.18$\pm$0.22 & -0.14$\pm$0.19 & \textbf{0.20$\pm$0.22} & \textbf{0.20$\pm$0.18} & 0.11$\pm$0.09 \\ \midrule
HUQ-DDU & \underline{0.73$\pm$0.11} & 0.19$\pm$0.24 & 0.07$\pm$0.29 & -0.09$\pm$0.21 & \textbf{0.21$\pm$0.09} \\
HUQ2-DDU & \underline{0.72$\pm$0.11} & \textbf{0.30$\pm$0.28} & \underline{0.11$\pm$0.30} & 0.00$\pm$0.28 & \underline{0.17$\pm$0.06} \\
HUQ-RDE & 0.68$\pm$0.21 & 0.16$\pm$0.33 & -0.01$\pm$0.25 & -0.05$\pm$0.22 & \underline{0.20$\pm$0.20} \\
HUQ2-RDE & \textbf{0.74$\pm$0.11} & \underline{0.26$\pm$0.31} & 0.09$\pm$0.27 & 0.01$\pm$0.34 & 0.10$\pm$0.22 \\
HUQ-MD & 0.58$\pm$0.24 & 0.02$\pm$0.31 & \underline{0.11$\pm$0.26} & \underline{0.02$\pm$0.26} & 0.12$\pm$0.09 \\
HUQ2-MD & 0.60$\pm$0.23 & 0.13$\pm$0.28 & 0.09$\pm$0.21 & -0.04$\pm$0.31 & 0.08$\pm$0.16 \\
\bottomrule
\end{tabular}
%}
\caption{\label{tab:psy_ue_rubioroberta_100}Results for the selective classification task for mental disorder detection datasets. The best results for each dataset are shown in bold. We underline top-3 methods after the best. The metric is normalized RC-AUC$\uparrow$ on the full curve.}
\end{table*}

\clearpage
\newpage

\section{Hyperparameter Values}
\label{appendix:hp}
  The optimal hyperparameters are presented in Table~\ref{tab:hp}. These hyperparameters are obtained using Bayesian optimization with early stopping. We train a model on the entire training dataset, use the validation dataset for metric evaluation, and select the optimal hyperparameters according to the best score on the validation set. We use the F1-micro score for the medical code prediction task, the F1 score for pathology class for the mental disorders detection task, and accuracy for others. We use the following hyperparameter grid:
  \begin{itemize}
    \item[\textit{ }] \textbf{Learning rate}: [5e-6, 6e-6, 7e-6, 9e-6, 1e-5, 2e-5, 3e-5, 5e-5, 7e-5, 1e-4];
  
    \item[\textit{ }]  \textbf{Num. of epochs for the medical code prediction task}: $\{n \in \mathbb{N} \mid 2 \leq n \leq 6\}$; 
  
    \item[\textit{ }] \textbf{Num. of epochs for other tasks}: $\{n \in \mathbb{N} \mid 2 \leq n \leq 15\}$;
  
    \item[\textit{ }] \textbf{Batch size}: [8, 16, 32, 64];

    \item[\textit{ }] \textbf{Batch size for mental disorders detection task}: [4, 8, 16, 32]; 
    
    \item[\textit{ }] \textbf{Weight decay}: [0, 1e-2, 1e-4]

    \item[\textit{ }] \textbf{Weight decay for mental disorders detection task}: [0, 1e-1, 1e-2]
  \end{itemize}

  \begin{table*}[htb!] \resizebox{\textwidth}{!}{\begin{tabular}{l|l|c|c|c|c|c}
\toprule
  \textbf{Model} &        \textbf{Dataset} &  \textbf{Accuracy Score} & \textbf{Learning Rate} &  \textbf{Num. Epochs} &  \textbf{Batch Size} &  \textbf{Weight Decay} \\
\midrule

GatorTron & MIMIC-III mortality & 0.840638 & 9e-6 & 4 & 16 & 0.01 \\
Clinical Longformer & MIMIC-IV MCP & 0.611338 & 9e-6 & 6 & 8 & 0 \\
Longformer & OV Dataset MCP & 0.724500 & 3e-5 & 3 & 16 & 0 \\
RuBioRoBERTa & DE & 0.809523 & 9e-6 & 9 & 4 & 0 \\
RuBioRoBERTa & DSM & 0.611111 & 5e-6 & 13 & 16 & 0.01 \\
RuBioRoBERTa & AL & 0.571428 & 6e-6 & 9 & 16 & 0.01 \\
RuBioRoBERTa & AD & 0.571428 & 2e-5 & 2 & 8 & 0.1 \\
RuBioRoBERTa & AC & 0.556962 & 1e-5 & 8 & 4 & 0 \\
\bottomrule
\end{tabular}
}\caption{\label{tab:hp} Optimal hyperparameters for each model and dataset.}\end{table*}

\end{document}